\documentclass[10pt,twocolumn,letterpaper]{article}

\usepackage{cvm}
\usepackage{times}
\usepackage{epsfig}
\usepackage{graphicx}
\usepackage{amsmath}
\usepackage{amssymb}
\usepackage{multirow}

\usepackage[pagebackref=true,breaklinks=true,letterpaper=true,colorlinks,bookmarks=false]{hyperref}

\newcommand{\keywords}[1]{{\bf \emph{Keywords: #1}}}

\cvmfinalcopy

\def\cvmPaperID{****} 

\ifcvmfinal\fi
\begin{document}

\def\eg{\emph{e.g.}} 
\def\Eg{\emph{E.g.}}
\def\ie{\emph{i.e.}} 
\def\Ie{\emph{I.e.}}
\def\cf{\emph{c.f.}} 
\def\Cf{\emph{C.f.}}
\def\etc{\emph{etc.}} 
\def\vs{\emph{v.s.}}
\def\wrt{w.r.t.} 
\def\dof{d.o.f.}
\def\etal{\emph{et al.}}

\title{Computer-Aided Layout Generation for Building Design: A Review}

\author{Jiachen Liu$^{1}$
\and
Yuan Xue$^{2}$
\and
Haomiao Ni$^{3}$
\and
Rui Yu$^{4}$
\and
Zihan Zhou$^{5}$
\and
Sharon X. Huang$^{1}$
\and
$^{1}$The Pennsylvania State University
\and
$^{2}$The Ohio State University
\and
$^{3}$University of Memphis
\and
$^{4}$University of Louisville
\and
$^{5}$Manycore Tech Inc.
}

\maketitle

\begin{abstract}
Generating realistic building layouts for automatic building design has been studied in both the computer vision and architecture domains. Traditional approaches from the architecture domain, which are based on optimization techniques or heuristic design guidelines, can synthesize desirable layouts, but usually require post-processing and involve human interaction in the design pipeline, making them costly and time-consuming. The advent of deep generative models has significantly improved the fidelity and diversity of the generated architecture layouts, reducing the workload by designers and making the process much more efficient. In this paper, we conduct a comprehensive review of three major research topics of architecture layout design and generation: floorplan layout generation, scene layout synthesis, and generation of some other formats of building layouts. For each topic, we present an overview of the leading paradigms, categorized either by research domains~(architecture or machine learning) or by user input conditions or constraints. We then introduce the commonly-adopted benchmark datasets that are used to verify the effectiveness of the methods, as well as the corresponding evaluation metrics. Finally, we identify the well-solved problems and limitations of existing approaches, then propose new perspectives as promising directions for future research in this important research area. A project associated with this survey to maintain the resources is available at 
\href{{https://github.com/jcliu0428/awesome-building-layout-generation}}{{awesome-building-layout-generation}}.\end{abstract}

\keywords{Computer-aided design, building layout, machine learning, deep generative models.}

\section{Introduction}

\begin{figure*}[!htb]
    \centering
    \scriptsize
    \setlength{\tabcolsep}{2pt}
    \begin{tabular}{ccc}

    \centering
        {\includegraphics[width=0.28\linewidth]{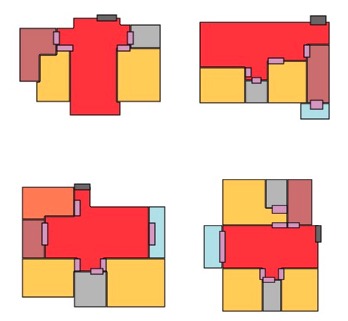}} &  
        {\includegraphics[width=0.28\linewidth]{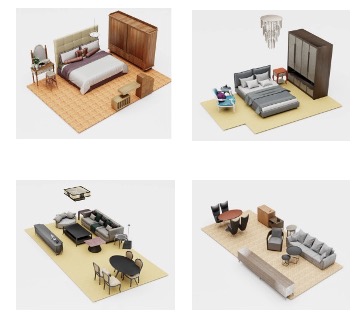}} &  
        {\includegraphics[width=0.28\linewidth]{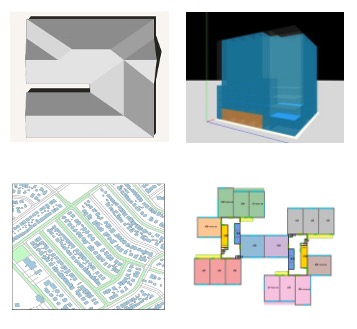}}\\

        Floorplan Layouts & Scene Furniture Layouts & Other Types of Building Layouts \\

    \end{tabular}
    \caption{The diverse building layout types used in mainstream computer-aided architectural design methods. The floorplan and scene furniture layouts are the two most common types of layouts. Moreover, there exist other types including roofs, building volumes, community or urban-level layout, etc.}
    \label{fig::teaser}
\end{figure*}


\par
Architecture design is important to ensure buildings and living spaces can fulfill their intended functions and requires careful consideration of layout, circulation, integration of amenities and furniture, as well as visual appearance. A well-designed residential house or community can provide a comfortable, salubrious and appealing environment for people to live in. However, the complete workflow of designers and architects usually involve handcrafted configurations from specific domains and multi-round refinement according to users' feedback, which makes this task time consuming and expensive. 

The advancement in image generation techniques~\cite{goodfellow2014generative, kingma2013auto, arjovsky2017towards,zhang2017stackgan, ho2020denoising} has inspired designers and researchers to rethink the workflow of architectural design. How to employ advanced machine learning techniques as assistance for computer-aided architectual design has captured attention and efforts from both architects and machine learning researchers. In the area of computer-aided building layout design, there are two mainstream research directions. The first one, residential floorplan layout generation, aims to synthesize diverse and realistic residential house layouts based on various types of user input to meet specific requirements. The second one, scene layout synthesis, endeavors to generate complete scenes including furniture objects that have desired semantic and spatial attributes such as object category, location, size and orientation. In addition to these two main topics, there are also other emerging topics such as block-plan generation, volumetric building design, roof synthesis, among others. Please refer to Fig.~\ref{fig::teaser} for several representative building layouts used in computer-aided design.

\par
Different from natural image synthesis, the problem of building layout generation has its distinct characteristics. First, the layout design process typically involves user interactions and constraints. A designer can specify the spatial connectivity, dimensions of rooms or furniture, according to customer needs, preferences, and building budget. 
This requires the layout generation model to have the ability to condition on user inputs as constraints, other than merely perform generation in an unconditional way. Second, unlike natural images which can be captured from arbitrary locations and viewpoints with irregular distributions, building layouts usually encapsulate notable structural regularities. For instance, the surrounding walls usually hold orthogonality~(also known as Manhattan-world assumption~\cite{coughlan2000manhattan}) for most buildings, and the walls are typically axis-aligned. For building designs, two apartments facing each other on the same floor often have similar or symmetric residential layouts. These regularities can serve as distribution priors which can be leveraged to alleviate the challenges associated with data sparsity, model complexity and parameter estimation. 

Last but not least, in natural image generation, a model with pixel-level encoding and decoding is usually expected to form an entire image, whereas in building layout generation, a set of vectors can usually appropriately represent the house or furniture layout. A rendering process which fills color into the generated vectorized boundary is followed for the purpose of displaying and visualization. These aforementioned key differences motivate researchers to develop diverse but distinctive design pipelines to address the layout generation problem.

\par
Considering the importance of computer-aided layout generation for building design, it is worthwhile to have a comprehensive survey to review existing methods and provide perspectives to inspire future research directions. A couple of related survey papers have been published. Weber~\etal~\cite{weber2022automated} gives a survey on automated floorplan generation paradigms, which are categorized into bottom-up, top-down methods from the architecture domain and referential methods from other domains especially machine learning~(ML). However, the scope of this survey is limited on architectural floorplan generation, rather than have a broader discussion on other types of building layout designs.
Ritchie~\etal~\cite{ritchie2023neurosymbolic} gives a comprehensive review of neurosymbolic models used in computer graphics. 
The paper focuses on a survey of representations for diverse geometric symbols and primitives, such as 2D or 3D shapes, materials and texture, in the architectural design area and beyond. 

These existing surveys either focus on a single topic (e.g. \cite{weber2022automated}) or cover topics that are too broad to be categorized as computer-aided layout generation~(such as \cite{ritchie2023neurosymbolic}).
To the best of our knowledge, there is no prior work that comprehensively provides a comprehensive introduction and overview on methods for computer-aided architecture design, specifically. 

In this paper, we aim to conduct a comprehensive review of methods for computer-aided layout generation for building design.  The layouts are not limited to floor plans, but also include scene layouts, building-level and site-level layouts. 

On each topic of interest, we first try to generally introduce the problem to handle, and formulate the task in a mathematical way. We then present 
existing approaches that could be categorized by various perspectives. 
We also summarize associated benchmark datasets used by current works as well as widely-adopted evaluation metrics to assess solutions for completing the task. 
At the end of the paper, we provide a summary of current research focuses and trends in this area, and present several perspectives related to addressing open challenges and initiating new research directions. 

\par
This survey paper is organized as follows: In Sec.~\ref{preliminary}, we give a brief introduction to the preliminaries of mainstream deep generative models from the machine learning and computer vision domains, which usually serve as the fundamental frameworks used by state-of-the-art~(SOTA) data-driven layout synthesis works. In Sec.~\ref{floorplan}, we formulate the task of residential floorplan generation, review methodologies as grouped by research domains or different formats of user input, then summarize widely-used datasets and evaluation metrics.
In Sec.~\ref{scene_synthesis}, we introduce the problem of scene layout synthesis and review the mainstream paradigms for tackling the problem, benchmarks datasets and evaluation protocols. In Sec.~\ref{other}, we review existing works for solving other problems related to building layout design such as building-level layout generation and site-level layout generation. In each section, we discuss the advantages and limitations of existing works, as well as the remaining challenges. 
Finally, in Sec.~\ref{discussions}, we present new perspectives that hopefully can inspire future research directions and outstanding works in this important area.

\section{Preliminaries of Deep Generative Models}

\begin{figure}[!htb]
    \centering
    \includegraphics[width=1.0\linewidth]{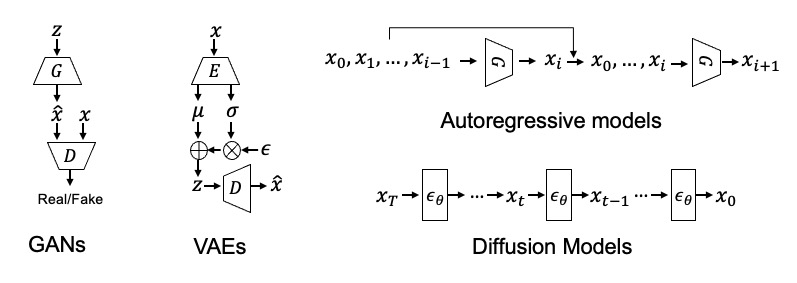}
    \caption{The typical workflow of different representative deep generative models, including GANs, VAES, Autoregressive Models and Diffusion Models~(DMs).}
    \label{fig::deep_generative_models}
\end{figure}

\label{preliminary}
Recent advancements in layout generation primarily leverage deep generative models, a class of techniques that employ deep neural networks to model the distribution of training samples \cite{bond2021deep}. We now present some commonly used deep generative modeling methods, such as generative adversarial networks (GANs), variational autoencoders (VAEs), autoregressive models, and the latest diffusion models (DMs). The typical workflow for each of these methods is intuitively illustrated in Fig.~\ref{fig::deep_generative_models}.

\subsection{Generative Adversarial Networks}
A typical Generative Adversarial Network (GAN) \cite{goodfellow2014generative} comprises two sub-networks: a discriminator network distinguishing between real and generated images, and a generator network creating images to deceive the discriminator. Specifically, with the input being a noise vector $z\sim p_\mathcal{Z}(z)$, the generator learns a differentiable mapping function $G(z)$ from noise space $\mathcal{Z}$ to data space $\mathcal{X}$. The discriminator, $D(x)$, is trained to output a single scalar indicating the probability that $x$ originates from real data rather than from the generator.
The discriminator $D$ is trained to maximize the probability of real data $x$ and minimize the probability of fake data generated by the generator $G$.
And $G$ is trained to minimize $\log(1-D(G(z)))$, thus maximizing $D(G(z))$ when $D$ is fixed. 
The overall training objective function can be defined as:
\begin{equation}
\label{vanilla_gan}
    \min_{G}\max_{D}\mathbb{E}_{x\sim p_\mathcal{X}(x)}[\log D(x)] + \mathbb{E}_{z\sim p_\mathcal{Z}(z)}[\log(1-D(G(z)))]\enspace.
\end{equation}

The training of GANs is widely recognized as challenging due to their inherent adversarial nature \cite{arjovsky2017towards}. As the discriminator becomes more effective, the gradients conveyed to the generator diminish; conversely, when the discriminator performs poorly, the generator does not receive informative gradients. Another issue is mode collapse, where one network becomes trapped in a bad local minimum, resulting in the learning of only a limited subset of the data distribution. To address these problems, many other loss functions have been proposed. One notable example is WGAN \cite{arjovsky2017wasserstein}, which utilizes the Wasserstein distance to measure the difference between distributions. The training objective function of WGAN is defined as:
\begin{equation}
\label{wgan}
    \min_{G}\max_{D}\mathbb{E}_{x\sim p_\mathcal{X}(x)}[D(x)] - \mathbb{E}_{z\sim p_\mathcal{Z}(z)}[D(G(z))]\enspace.
\end{equation}
Note that unlike the original GAN model Eq.~\eqref{vanilla_gan}, where the discriminator $D$ serves as a binary classifier, in WGAN, $D$ is employed to approximate the Wasserstein distance, which is a regression task. Consequently, the sigmoid function, usually present in the last layer of $D$, is omitted in WGAN. WGAN also uses weight clipping as a strategy to enable network training with the approximated Wasserstein distance.

\subsection{Variational Autoencoders}
Consider a latent-based model $p_\theta(x|z)$ with a prior $p_\theta(z)$ and a posterior $p_\theta(z|x)$. 
As the direct computation of $p_\theta(x)=\int_zp_\theta(x|z)p_\theta(z)dz$ is not tractable, 
to maximize the data likelihood $p_\theta(x)$, 
one can leverage variational inference \cite{kingma2013auto} to establish a tractable lower bound on $p_\theta(x)$. This can be achieved by introducing an approximation of the true intractable posterior, denoted as $q_\phi(z|x)=\text{argmin}_qD_{KL}(q_\phi(z|x)||p_\theta(z|x))$. In particular, variational autoencoders employ a feed-forward inference network to approximate $q_\phi(z|x)$, enabling scalability to large datasets \cite{kingma2013auto,rezende2014stochastic}. From the definition of KL divergence \cite{csiszar1975divergence}, we have:
\begin{equation}
    \label{eq:kl}
    D_{KL}(q_\phi(z|x)||p_\theta(z|x)) = \mathbb{E}_{q_\phi(z|x)}\left[\log\frac{q_\phi(z|x)}{p_\theta(z|x)}\right]\enspace.
\end{equation}
Thus we can get:
\begin{align}
\begin{split}
    \label{eq:elbo}
    \log p_\theta(x) =&\enspace D_{KL}(q_\phi(z|x)||p_\theta(z|x)) - \mathbb{E}_{q_\phi(z|x)}[\log q_\phi(z|x)] \\ & + \mathbb{E}_{q_\phi(z|x)}[\log p_\theta(z, x)] \\
    \geq&\enspace -\mathbb{E}_{q_\phi(z|x)}[\log q_\phi(z|x)] + \mathbb{E}_{q_\phi(z|x)}[\log p_\theta(z, x)] \\
    =&\enspace -\mathbb{E}_{q_\phi(z|x)}[\log q_\phi(z|x)] + \mathbb{E}_{q_\phi(z|x)}[\log p_\theta(z)] \\ & + \mathbb{E}_{q_\phi(z|x)}[\log p_\theta(x|z)] \\
    =&\enspace -D_{KL}(q_\phi(z|x)||p_\theta(z)) + \mathbb{E}_{q_\phi(z|x)}[\log p_\theta(x|z)] \\
    \equiv&\enspace \mathcal{L}(\theta, \phi; x)\enspace,
\end{split}
\end{align}
where $\mathcal{L}$ is known as the evidence lower bound (ELBO) \cite{wainwright2008introduction}. Hence, maximizing the data likelihood estimation can be attained by maximizing the ELBO $\mathcal{L}$ in Eq.~\ref{eq:elbo}. To optimize this bound with respect to parameters $\theta$ and $\phi$, gradients need to be back-propagated through the stochastic sampling process $z\sim q_\phi(z|x)$. This can be achieved by reparameterizing $z$ to move the sampling operation to an input layer. Specifically, when $z\sim q_\phi(z|x)=\mathcal{N}(z;\mu, \sigma^2\mathbf{I})$, we can sample $z$ by first sampling $\epsilon\sim \mathcal{N}(0, \mathbf{I})$, and then computing $z=\mu+\sigma\cdot\epsilon$. 

\subsection{Autoregressive Models}
Based on the chain rule of probability, autoregressive models~\cite{bengio2000neural} generate the variables in an iterative manner. Suppose the variable to be generated can be decomposed as \textbf{x} = $x_{1}, x_{2}, ..., x_{n}$, the probability of generating such a sequence can be represented as:
\begin{equation}
    p(x) = p(x_{1}, x_{2}, ..., x_{n}) = \prod_{i=1}^{n}p(x_{i}|x_{1}, x_{2}, ..., x_{i-1}).
\end{equation}
In practice, to make the optimization process easier for the network, the negative log-likelihood $-ln~p(x)$ is minimized during training, which is essentially equal to maximizing the original probability $p(x)$:

\begin{equation}
    -ln~p(x) = - \sum_{i=1}^{n}~ln~p(x_{i}|x_{1}, x_{2}, ..., x_{i-1}).
\end{equation}

The common network architectures of autoregressive models include masked MLPs~\cite{larochelle2011neural}, recurrent neural networks~(RNNs) such as LSTMs~\cite{HochSchm97} and GRUs~\cite{chung2014empirical}, causal convolutions~\cite{chen2018pixelsnail} and masked self-attention-based Transformers~\cite{vaswani2017attention}.

Autoregressive models are natural choices when the modalities for generation can be represented as ordered sequences, such as text or audio.
For other modalities such as images, one would need to find a way to transform data in the original modalities into sequence representations. A common limitation of autoregressive models is that when the sequence turns longer, the computational cost increases drastically, leading to much slower generation for the later time steps of the sequence. On layout generation problems, since layouts can be typically represented as vectorized sequences, autoregressive models, known for their effectiveness in this representation capacity on sequential data, are widely adopted to model the generation process. We will discuss the details of such works in the next sections.

\subsection{Diffusion Models}
Diffusion models (DMs) \cite{ho2020denoising,sohl2015deep,song2019generative} are probabilistic models designed to
learn a data distribution. DMs typically consist of a forward process and a reverse process. 
Given a sample from the data distribution $x_0\sim q(x_0)$, the DM \textit{forward} process produces a Markov chain $x_1, \dots, x_T$ by gradually adding Gaussian noise to $x_0$ based on a variance schedule $\beta_1, \dots, \beta_T$, that is:
\begin{equation}
\label{eq:forward}
    q(x_t|x_{t-1}) = \mathcal{N}(x_t; \sqrt{1-\beta_t}x_{t-1}, \beta_t\mathbf{I})
\enspace,
\end{equation}
where variances $\beta_t$ are constants.
If $\beta_t$ are small, the posterior $q(x_{t-1}|x_{t})$ can be well approximated by diagonal Gaussian \cite{nichol2021glide,sohl2015deep}. Furthermore, when the $T$ of the chain is large enough, $x_T$ can be well approximated by standard Gaussian distribution $\mathcal{N}(\mathbf{0}, \mathbf{I})$. These suggest that the true posterior $q(x_{t-1}|x_{t})$ can be estimated by $p_\theta(x_{t-1}|x_t)$ defined as \cite{nichol2021improved}:
\begin{equation}
\label{eq:reverse}
    p_\theta(x_{t-1}|x_t)=\mathcal{N}(x_{t-1}; \mu_\theta(x_t), \Sigma_\theta(x_t))
\enspace.
\end{equation}
The DM \textit{reverse} process (also known as \textit{sampling}) then generates samples $x_0\sim p_\theta(x_0)$ by initiating a Markov chain with Gaussian noise $x_T\sim \mathcal{N}(\mathbf{0}, \mathbf{I})$ and progressively decreasing noise in the chain of $x_{T-1}, x_{T-2}, \dots, x_0$ using the learnt $p_\theta(x_{t-1}|x_t)$. To learn $p_\theta(x_{t-1}|x_t)$, Gaussian noise $\epsilon$ is added to $x_0$ to generate samples $x_t\sim q(x_t|x_0)$, then a model $\epsilon_\theta$ is trained to predict $\epsilon$ using the following mean-squared error loss:
\begin{equation}
\label{eq:dm}
    L_\text{DM}=\mathbb{E}_{t\sim \mathcal{U}(1, T), x_0\sim q(x_0), \epsilon\sim \mathcal{N}(\mathbf{0}, \mathbf{I})}[||\epsilon-\epsilon_\theta(x_t, t)||^2]
\enspace,
\end{equation}
where time step $t$ is uniformly sampled from $\{1, \dots, T\}$. Then $\mu_\theta(x_t)$ and $\Sigma_\theta(x_t)$ in Eq.~\eqref{eq:reverse} can be derived from $\epsilon_\theta(x_t, t)$ to model $p_\theta(x_{t-1}|x_t)$ \cite{ho2020denoising,nichol2021improved}. The denoising model $\epsilon_\theta$ is typically implemented using a time-conditioned U-Net \cite{ronneberger2015u} with residual blocks \cite{he2016deep} and self-attention layers \cite{vaswani2017attention}. Sinusoidal position embedding \cite{vaswani2017attention} is also usually used to specify the time step $t$ to $\epsilon_\theta$.

\begin{figure*}[!htb]
    \centering
    \includegraphics[width=0.9\linewidth]{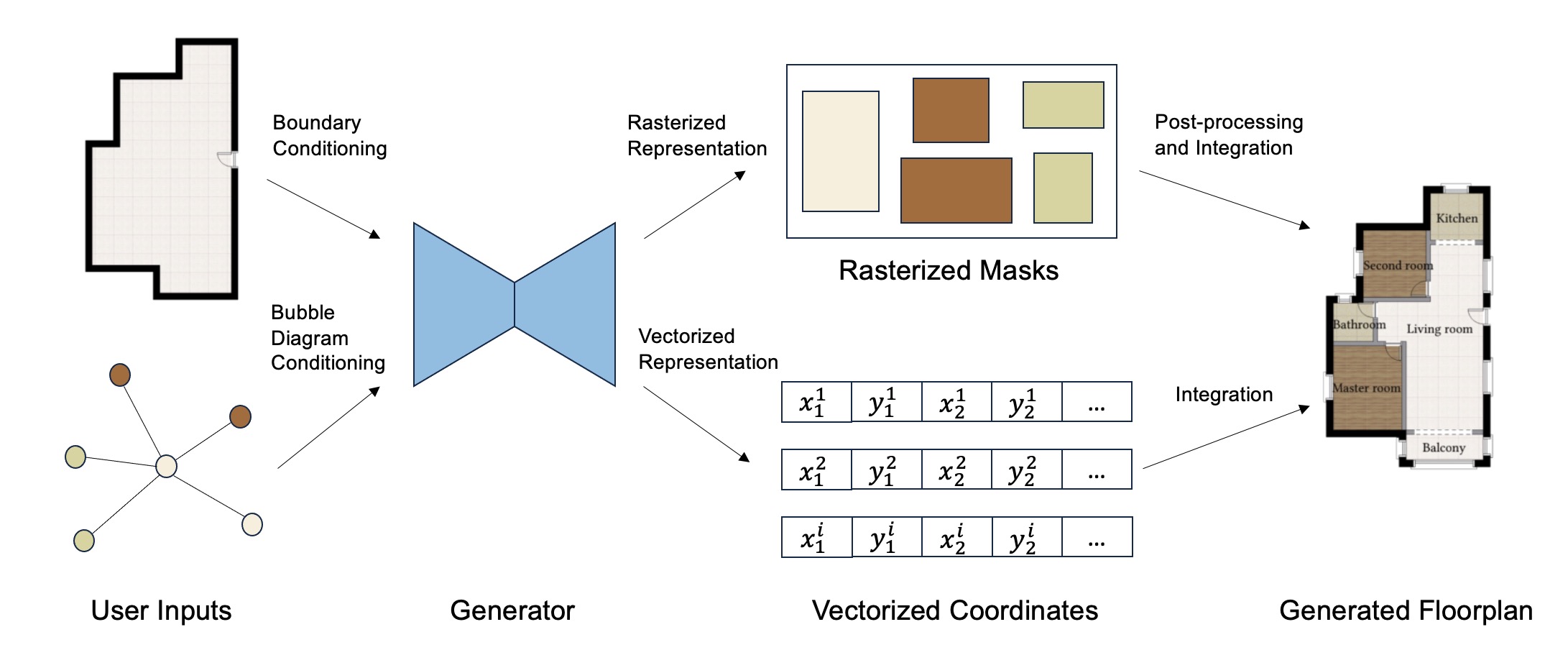}
    \caption{A generic learning-based floorplan generation pipeline with user input. We show two major user input conditions, residential boundary and bubble diagram--commonly used in current studies. Methods using rasterized representation aim to generate a set of room masks then perform post-processing and integrate the outputs into a vectorized floorplan. For vectorized methods, the outcomes can be directly integrated into a floorplan. The icons of the boundary input and the final generation are referred from the diagrams used in RPLAN.}
    \label{fig::floorplan_pipeline}
\end{figure*}

\section{Residential Floorplan Layout Generation}
\label{floorplan}
In this section, we start with the task of interior floorplan layout generation, which is a representative layout format among all types of building layouts.
In terms of methodology, we categorize the approaches of the generation task into two groups: traditional optimization-based approaches from the architectural domain, and data-driven learning-based approaches from the domains of machine learning and computer-aided design. We review methods that are recently proposed in each group. 
We also provide a summary of publicly-available benchmark datasets and widely-adopted evaluation protocols for the task. The section ends with a comparative analysis of competing methods for the task. 

\subsection{Task formulation}
\label{floorplan_formulation}

A floorplan contains both room elements as well as connection elements such as doors and windows. For simplicity, and unless specified otherwise, the remainder of this survey paper will use the term `room' to refer to all these elements, without distinguishing between `room' and `connection'. 
Because of the geometric attributes of real-world room layouts, the vertices of a room $R$ are usually represented as a polygonal vector which consists of a group of self-looped corner coordinates:
\begin{equation}
\boldsymbol{R} = \{\boldsymbol{r}_{x_{i}}, \boldsymbol{r}_{y_{i}}\}_{i=1}^{n}
\end{equation}
where $r_{x_{i}}$ and $r_{y_{i}}$ are the $x$ and $y$ coordinates of the $i$-th corner of the room $R$, and 
$n$ is the total number of room corners. Please note that $n$ can be different for each room. 
Stacking all the rooms together as a sequence, the entire floorplan $L$ for a residential house can be represented as:
\begin{equation}
\begin{aligned}\label{eqn:seq}
L = \{\boldsymbol{R}_{j}\}_{j=1}^{N}.
\end{aligned}
\end{equation}
where $N$ is the number of rooms. $N$ can be different for each floorplan sample.
\par
Please note that typical floorplan layouts satisfy the Manhattan-world constraint, \ie, their boundaries are axis aligned. Some recent study~\cite{shabani2023housediffusion} has overcome such limitation and proposed a general representation where non-Manhattan layouts can also be synthesized. The generation process is typically constrained by some specified user input or conditions such as house boundary $F$, and bubble diagram $G$~(which is a graph indicating the room types and their connectivity relationships). More recently, a floorplan restoration task is proposed~\cite{hosseinifloorplan}, which aims to complete and recover the entire floorplan $L$, from a partial reconstruction outcome $L_{p}$. Thus, without loss of generality, the generation process, which takes some of the conditions as input, can be formulated as:
\begin{equation}
L = M(F, G, L_{p}),
\end{equation}
where $M$ denotes the floorplan generative model.

\subsection{Traditional Methods from Architectural Domain}

\begin{table*}[!htb]
\caption{Learning-based floorplan generation methods with their floorplan representation, the types of user input, utilized benchmark datasets and publishing venue. }
\resizebox{1.0\linewidth}{!}{
\begin{tabular}{l|c|c|c|c|c}
\hline
{Methods} & {Representation} & {Framework} & {User Input} & {Benchmark Datasets} & {Publishing Press}\\
\hline
Wu~\etal & Rasterized & CNNs & House Boundary & RPLAN & SIGGRAPH Asia 2019\\
Graph2Plan & Rasterized & CNNs, GNNs & House Boundary, Bubble Diagram & RPLAN & SIGGRAPH 2020\\
HouseGAN & Rasterized & GANs & Bubble Diagram & RPLAN, LIFULL & ECCV 2020\\
HouseGAN++ & Rasterized & GANs & Bubble Diagram & RPLAN & ICCV 2021 \\
Para~\etal & Vectorized & Transformers & Unconditioned & RPLAN, LIFULL & ICCV 2021\\
iPLAN & Rasterized & GANs, CNNs & House Boundary & RPLAN, LIFULL & CVPR 2022\\
Liu~\etal   & Vectorized & GNNs, Transformers & Bubble Diagram & RPLAN & ECCV 2022\\
Upadhyay\etal & Rasterized & Conv-MPNs, CNNs & House Boundary, Bubble Diagram & RPLAN & ICMEW 2022\\
FloorplanGAN & Vectorized & GANs & Room types and areas & RPLAN & Automation in Construction 2022\\
HouseDiffusion & Vectorized & Diffusion Models & Bubble Diagram & RPLAN & CVPR 2023\\
Tang~\etal & Rasterized & GANs & Bubble Diagram & LIFULL & CVPR 2023\\
Zheng~\etal & Vectorized & GNNs & Bubble Diagram & RPLAN & Automation in Construction 2023\\
Aalaei~\etal & Rasterized & Conv-MPNs, GANs & Bubble Diagram & RPLAN & Automation in Construction 2023\\
Hosseini~\etal & Rasterized & CNNs, Transformers & Partial Floorplan & RPLAN and a new restoration dataset & BMVC 2023\\
WallPlan & Vectorized & CNNs & House Boundary & RPLAN & ACM Transactions on Graphics (TOG) 2022\\
MaskPlan & Vectorized & Masked Autoencoders, Transformers & House Boundary and Partial Attributes & RPLAN & CVPR 2024\\
Hu~\etal & Vectorized &  Transformers, Diffusion Models & House Boundary, Bubble Diagram & RPLAN & Arxiv 2024\\
\hline
\end{tabular}
}
\centering
\label{tab:floorplan_methods}
\end{table*}

Traditional floorplan generation methods from the architecture design domain are typically categorized as either bottom-up or top-down approaches~\cite{fan2023automated}, each with their own strengths and weaknesses. We introduce some representative works for each type in this subsection. 
\paragraph{Bottom-up methods} Building designs are usually constrained by specified spatial requirements, such as room dimensions and mutual adjacency. Therefore, bottom-up paradigms tend to become a natural choice for mapping spatial relations or bubble diagrams. Conceptually, a group of predefined building components are aggregated into a larger assembly following specific regulations and constraints. Rosenman \etal~\cite{rosenman2000case} design a single floorplan generation pipeline that maximizes cross ventilation and minimizes the weighted sum of room distances. \cite{arvin2002modeling} describes how to apply topological and geometric objectives to house design within proper boundaries. Merrell \etal~\cite{merrell2010computer} showcases how to accommodate complex input room sequences into a suitable residential layout through a Bayesian network on multi-floor buildings. Beyond spatial constraints, Yi \etal~\cite{yi2014performance} propose designing 3D house layouts based on optimal environmental performance. Guo \etal~\cite{guo2017evolutionary} design a multi-agent topology-finding system and an evolutionary optimization process to first generate topology-satisfied layouts and then achieve predefined architectural criteria. GPLAN \cite{shekhawat2020gplan} employs graph-theoretical and optimization techniques to facilitate the generation of dimensioned floorplan layouts.
\paragraph{Top-down methods} Top-down methods are inspired by predefined building massing from urban-scale considerations, with strong constraints on the envelope in real-world architectural design. As a result, subdivision, fitting, shape packing, and iterative agent-based methods have been proposed and employed to automate design problems across various architectural scales. Top-down methods take a massing or boundary as input and a set of entities as targets for insertion. The input is then subdivided based on geometric constraints to assign spaces. Compared to bottom-up methods, which require a specified optimization process to ensure boundary constraints, top-down methods naturally conform to this condition by subdividing and transforming directly from the global boundary. Medjdoub \etal~\cite{medjdoub2000separating} design a subdivision pipeline for single floorplan generation, considering adjacency and room scale constraints. Banerjee \etal~\cite{banerjee2008model} propose a computational model for creative design, highlighting practicality, originality, and support for interactive user input. \cite{feng2016crowd} presents a mid-scale floorplan layout generation approach by optimizing with respect to human crowd properties, including mobility, accessibility, and coziness, using agent-based crowd simulation. For a more extensive review of traditional methods proposed in the architectural domain, please refer to \cite{fan2023automated}. 

\begin{figure}[!htb]
    \centering
    \includegraphics[width=1.0\linewidth]{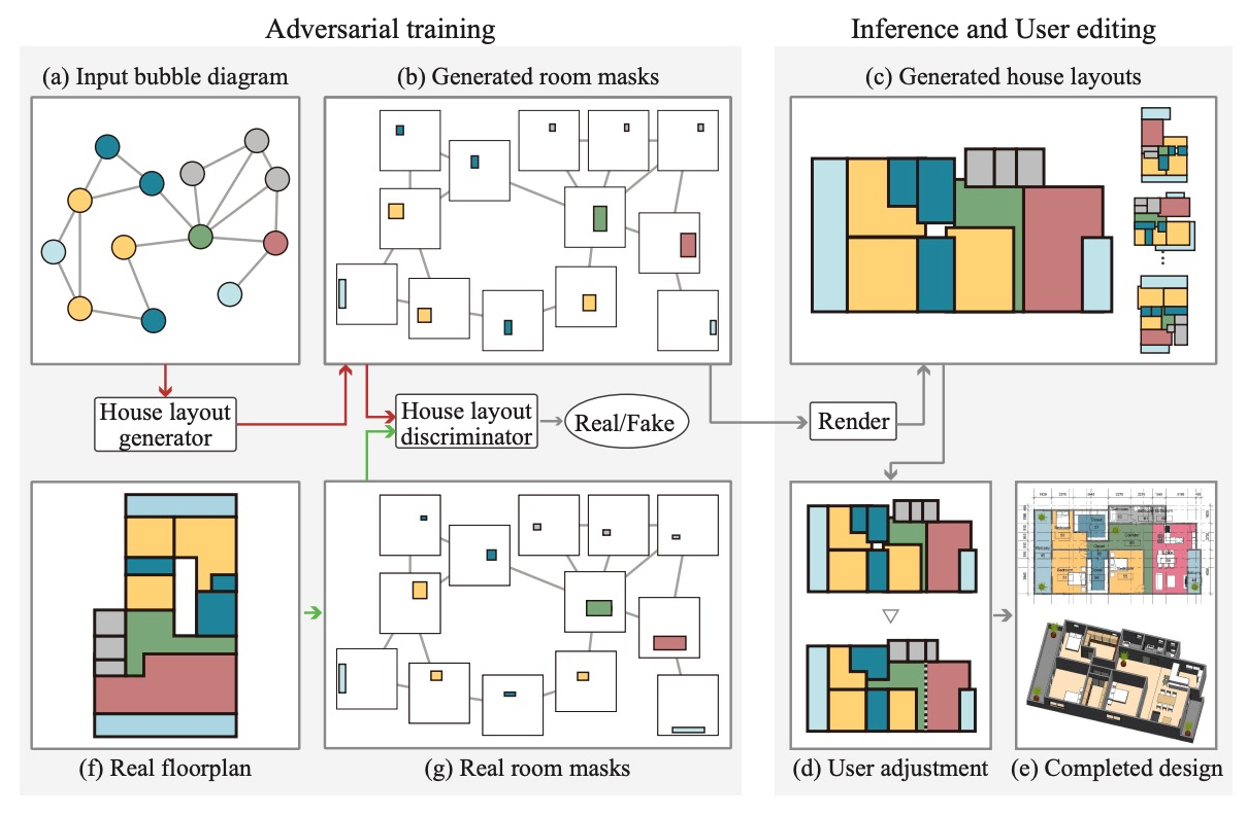}
    \caption{HouseGAN is a representative floorplan generation approach using the rasterized representation. It first parses the input bubble diagram, then generates the room masks separately with a generator and discriminator architecture. The separately generated rooms are subsequently integrated together and post-processed to finalize the floorplan design.}
    \label{fig::housegan}
\end{figure}

\begin{figure}[!htb]
    \centering
    \includegraphics[width=1.0\linewidth]{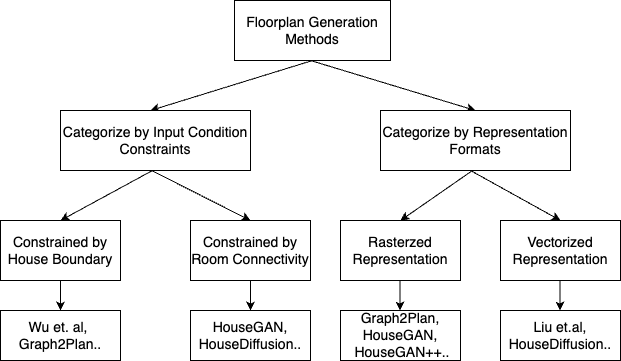}
    \caption{A tree-structured diagram to illustrate the categorization of different approaches and representative methods.}
    \label{fig::floorplan_treenode}
\end{figure}

\subsection{Data-Driven Learning-based Methods}

In this subsection, we review learning-based approaches for computer-aided layout generation proposed in the machine learning and computer-aided design domains. We group the methods based on the utilized representation format and the user input conditions. A generic pipeline of the methods with user-input is demonstrated in Fig.~\ref{fig::floorplan_pipeline}, and a tree-structured diagram on how we categorize approaches is shown in Fig.~\ref{fig::floorplan_treenode}.

\subsubsection{Methods Categorized by Layout Representation Format}

We focus on floorplan layouts and first categorize the deep generative methods by how they represent floorplans during generation. In general, there are two groups of methods, those using a rasterized representation and those using a vectorized representation. Methods using the rasterized representation~\cite{hu2020graph2plan, nauata2020house, nauata2021house, upadhyay2022flnet} treat the rooms or walls of the floorplan as rasterized masks, then composite the rooms into a floorplan image. Methods using the vectorized representation~\cite{para2021generative, liu2022end, luo2022floorplangan, shabani2023housediffusion} represent the room coordinates as quantized vectors, enabling a more straightforward and efficient representation of a floorplan layout.

\paragraph{Methods using Rasterized Representation.}

The most straightforward way to represent floorplans is to treat them similarly to natural images. Each room is represented as a rasterized mask, and the image pixels are the generation targets. RPLAN~\cite{wu2019data} and Graph2Plan~\cite{hu2020graph2plan} adopt convolutional neural networks (CNNs) and graph neural networks (GNNs) to generate graph-constrained rasterized floorplans, which are then converted into floorplan vectors via post-processing. Leveraging GAN-based models, HouseGAN~\cite{nauata2020house}~(shown in Fig.~\ref{fig::housegan}) learns to generate a list of rasterized masks for rooms using a bubble diagram as input. HouseGAN++\cite{nauata2021house} extends HouseGAN by introducing a GT-conditioning training scheme and a set of test-time optimization strategies. iPLAN\cite{he2022iplan} proposes a human-in-the-loop system that enables human experts and the deep learning framework to co-evolve a sketchy layout into the final floorplan design, generating segmented room masks in the process. Upadhyay \etal~\cite{upadhyay2022flnet} take both bubble diagrams and input boundaries as conditions and process them by different feature embedding networks. The bubble diagram embeddings and boundary features are then aggregated via a cascaded alignment network to generate the final floorplan layout. Tang \etal~\cite{tang2023graph} propose a graph transformer architecture coupled with GANs to generate floorplans in an adversarial procedure. For these methods, handcrafted post-processing strategies are usually adopted to convert the masks into vectors as the final representation. Aalaei \etal~\cite{aalaei2023architectural} utilize a Conv-MPN network to parse the bubble diagram and a GAN architecture to generate rasterized floorplans. The generation proceeds iteratively until the geometrical and topological constraints are approximately satisfied, after which the final output is transformed into a vectorized format.

\begin{figure}[!htb]
    \centering
    \includegraphics[width=1.0\linewidth]{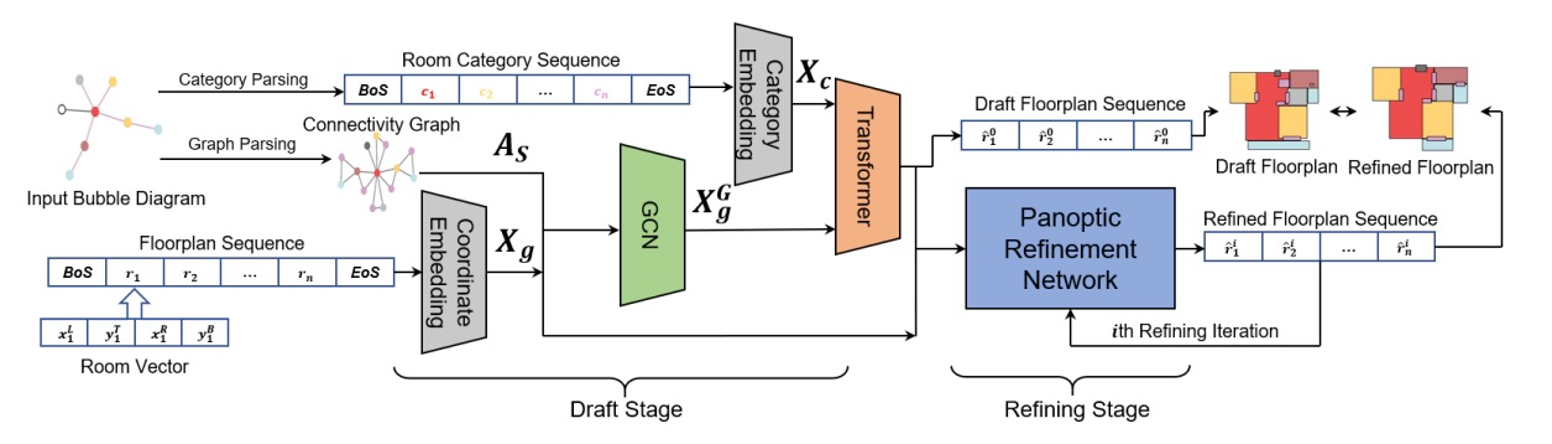}
    \caption{Liu~\etal represents the floorplan samples as vectors and presents an end-to-end vectorized floorplan generation pipeline. In the first stage, the GCN and Transformer-based network takes the parsed bubble diagram embeddings as input and generates a draft floorplan sequence in an auto-regressive manner. In the second stage, the draft floorplan is passed to another Transformer network to achieve panoptic refinement for the synthesized floorplan.}
    \label{fig::panopticrefine}
\end{figure}

\paragraph{Methods using Vectorized Representation.}

Instead of representing rooms as rasterized segments, methods using the vectorized representation~\cite{para2021generative, liu2022end, shabani2023housediffusion} encode room coordinates as 1-D vectors. By utilizing 1-D vectors, this representation allows for improved generation efficiency while preserving the topological and geometric integrity of the layout.
Para \etal~\cite{para2021generative} use an autoregressive Transformer model primarily for unconditioned floorplan layout generation. Liu \etal~\cite{liu2022end} propose a two-stage vectorized floorplan generation framework conditioned on bubble diagram input. The first stage generates a draft floorplan in an autoregressive manner, and the second stage refines the draft floorplan with panoptic refinement (see Fig. \ref{fig::panopticrefine}). FloorplanGAN~\cite{luo2022floorplangan} takes the room types and their relative occupancy area as input, then extracts room type and geometric embeddings from these conditions. A GAN-based architecture is then used for floorplan generation, consisting of a vector generator, a differentiable renderer, and a CNN-based discriminator. More recently, HouseDiffusion~\cite{shabani2023housediffusion} employs diffusion models~\cite{ho2020denoising, song2020denoising} to represent and generate floorplan vectors. The framework is trained through a diffusing-then-denoising process, and during inference, the final generated floorplan is iteratively denoised from Gaussian noise, conditioned on the input bubble diagram. A significant contribution of their work is breaking the Manhattan-layout assumption commonly adopted in existing works~\cite{para2021generative, liu2022end}, allowing their proposed diffusion model to generate diverse non-Manhattan floorplans. Zheng \etal~\cite{zheng2023neural} first parse a dual graph structure from the given bubble diagram, then apply a hybrid GNN along with a topology and geometry-driven optimization approach to generate room coordinates. MaskPlan~\cite{zhang2024maskplan} employs a Masked Autoencoder architecture to generate a blend of graph-based and image-based layout attributes, conditioning on partial input formats from users. Compared to rasterized representation, vectorized representation is a more desirable choice in recent studies. Representing rooms with varying sizes using 1D vectors~(as described using the format in Sec.~\ref{floorplan_formulation}) significantly enhances generation efficiency and improves the ability to preserve correct floorplan topology and geometry. Constraints can be naturally integrated within the vectorized format. For example, the number of rooms in a generated layout can be controlled by specifying the length of the vectors, ensuring compliance with design specifications while maintaining structural consistency. Moreover, vectorized coordinates inherently support axis-aligned properties more effectively than rasterized formats, ensuring precise spatial alignment. 
Vectorized representation also enables end-to-end optimization explicitly on the graphic vectors, eliminating the need for post-processing steps to transform rasterized masks into vectors. 

\subsubsection{Methods Categorized by User Input Conditions}

Floorplan design is typically a conditional generation problem that takes various conditions from users or designers as constraints. In this subsection, we discuss the common user input conditions and categorize them into two mainstream formats: residential boundary constraints and bubble-diagram-based constraints. We then introduce the deep generative methods associated with each respective category.

\paragraph{Methods Conditioned on Residential Boundary Constraints.}
Since residential houses are often not independent and share common building primitives, and because design budget is an important factor, the residential boundary is a common input constraint for floorplan synthesis, specified by either designers or users. In RPLAN~\cite{wu2019data}, a CNN network takes the rasterized house boundary as input and predicts room types, locations, and rasterized walls. Graph2Plan~\cite{hu2020graph2plan} takes the boundary image along with the retrieved room graph from users as input, then encodes this information through Graph Neural Networks (GNN~\cite{scarselli2008graph}) and CNN networks. iPLAN~\cite{he2022iplan} takes the apartment boundary and room types as conditions, and learns to locate and segment the rooms into rasterized masks that conform to the boundary. WallPlan~\cite{sun2022wallplan} takes the boundary as input, initializes a network to learn window locations, and then passes the window-conditioned house boundary into parallel branches to learn wall graphs and room semantics. Most recently, Hu~\etal~\cite{hu2024advancing} proposes a transformer-based diffusion model for wall junction generation and wall segment prediction, with a focus on enforcing the geometric attributes from structured graphs.

\paragraph{Methods Conditioned on Bubble Diagram Constraints.}
Another line of work~\cite{nauata2020house, nauata2021house, liu2022end, shabani2023housediffusion} uses another common input format known as the bubble diagram, which includes room types and graph connectivities, as input constraints. For room types, they are either encoded as one-hot vectors in rasterized methods~\cite{nauata2020house, nauata2021house} or quantized as tokens and encoded through learnable embeddings in vectorized representations~\cite{liu2022end}. Message passing networks~\cite{gilmer2017neural} or GNNs~\cite{scarselli2008graph, kipf2016semi} are employed in~\cite{nauata2020house, nauata2021house, liu2022end, upadhyay2022flnet} to encode the element connectivity between different rooms. Tang~\etal~\cite{tang2023graph} designs a graph-constrained transformer framework to learn graph-based relations from the input bubble diagrams. HouseDiffusion~\cite{shabani2023housediffusion} encodes the graph-constrained room relationship via performing graph-aware attention on the embedded room coordinates using the given bubble diagram. Zheng~\etal~\cite{zheng2023neural} also leverages the bubble diagram input to parse the dual graph representation to guide the optimization stage. Aalaei~\etal~\cite{aalaei2023architectural} uses Conv-MPN~\cite{zhang2020conv} to exploit topology information from the input bubble diagram and yields floorplans through a GAN architecture.

\subsection{Benchmark datasets}
\paragraph{RPLAN~\cite{wu2019data}} RPLAN is the first large-scale
densely annotated floorplan dataset with over $80K$ real residential buildings, which is widely adopted in ML-based floorplan generation works. It provides comprehensive vectorized graphics of room types, boundaries, dimensions, and connectivity information.
\paragraph{LIFULL~\cite{weko_3670_1}} LIFULL HOME is a large-scale database containing millions of real floorplan samples with corresponding room type labels. The bubble diagram can be retrieved if needed through a floorplan vectorization~\cite{liu2017raster} process.
\paragraph{Structured3D~\cite{zheng2020structured3d}} Structured3D is a photo-realistic dataset with diverse 3D structure annotations, satisfying a range of 3D modeling problems. It provides scene information from both holistic views such as floorplans and 3D meshes, and primitives such as junctions, lines and planes. Floorplans are annotated by room types and dimensions over 3,500 scenes.
\paragraph{Zillow~\cite{cruz2021zillow}} The Zillow Indoor Dataset (ZInD) is a large-scale real indoor dataset with over $70k$ panoramas and unfurnished houses. Compared to other benchmarks, it provides more diverse data following real-world distribution, containing both Manhattan and non-Manhattan layouts. The floorplans are annotated with room semantics and dimensions over 2,500 scenes.
\paragraph{CubiCasa5k~\cite{kalervo2019cubicasa5k}} Cubicasa5K is a floorplan dataset containing 5,000 sampled residential floorplans. A unique feature of this dataset is that it provides furniture object-level vectorized layouts, enabling more fine-grained intelligent design for both room and furniture layouts.
\paragraph{ProcTHOR-10k~\cite{deitke2022}} PROCTHOR is a framework for procedural generation of Embodied AI environments. PROCTHOR produces a large and diverse set of floorplans, followed by a large asset library of 108 object types and 1633 fully interactable instances that are used to automatically populate each floorplan. 
\paragraph{DStruct2Design~\cite{luo2024dstruct2design}} In the DStruct2Design paper, the authors merge existing datasets, including RPLAN and ProcTHOR, and apply post-processing to format them for compatibility with language-based generative models. This enables language-guided floorplan generation that can specify both geometric structures and the absolute dimensions of rooms.

\begin{table*}[htb]
\caption{Quantitative comparison on SOTA bubble-diagram-constrained floorplan generative methods. The numbers~(5, 6, 7, 8) or `mixed' refer to the room number of the floorplans used in the testing split.}
\centering
\resizebox{1.0\linewidth}{!}{
\setlength{\tabcolsep}{4pt}{
\begin{tabular}{l|ccccc|ccccc|c|c}
\hline
 \multirow{2}{*}{Methods} & \multicolumn{5}{c|}{Diversity~$\downarrow$} & \multicolumn{5}{c|}{Compatibility~$\downarrow$} & \multicolumn{1}{c|}{Inference Memory~(GB) $\downarrow$} & \multicolumn{1}{c}{Latency~(FPS) $\uparrow$}\\
& Mixed & $5$ & $6$ & $7$ & $8$ & Mixed & $5$ & $6$ & $7$ & $8$ & $8$ & $8$ \\
\hline 
HouseGAN++ & 17.9/0.2 & 19.9/0.3 & 15.4/0.1 & 14.0/0.2 & 18.9/0.5 & 2.7/0.1 & 1.7/0.0 & 2.1/0.0 & 3.1/0.1 & 3.6/0.1 & 1.6 & 2.83\\
PanopticRefine & 16.3/0.4 & 18.9/0.5 & 16.9/0.5 & 14.5/0.3 & 16.5/0.5 & 2.5/0.1 & 1.3/0.0 & 1.9/0.0 & 3.4/0.0 & 5.0/0.0 & 1.27 & 4.22\\
HouseDiffusion & 10.5/0.2 & 11.2/0.2 & 10.3/0.2 & 10.4/0.2 & 9.5/0.1 & 2.0/0.0 & 1.5/0.0 & 1.2/0.0 & 1.7/0.0 & 2.5/0.0 & 1.9 & 1.34 \\
\hline
\end{tabular}
}}
\label{tab:quantitative_floorplan}
\end{table*}

\subsection{Evaluation metrics and experimental comparison}
In this section, we introduce common evaluation metrics for conditioned floorplan generation. Tab.~\ref{tab:quantitative_floorplan} illustrates the evaluation metrics used in representative vectorized methods, including \textit{diversity and compatibility}. Additionally, \textit{realism} measures the generation quality by user engagement and ranking on different methods, which is another important metric.

\paragraph{Realism}  Realism is one of the most important metrics in generation tasks, as it evaluates user perception of the generated floorplans. A common practice is to invite a group of participants—either professional architects or amateurs—to provide scores or rankings for the outcomes of various generation methods, given the same set of sampled constraints. The average score or ranking is then calculated to assess the quality of the generation.
\paragraph{Diversity and Quality} 
 The Fréchet Inception Distance (FID)\cite{heusel2017gans} is a widely used metric for assessing the quality of generated images by comparing their distribution to that of real images. It calculates the Fréchet distance between the feature representations of generated and real images, which are extracted using a pre-trained Inception network, measuring differences in both the mean and covariance. Lower FID scores indicate that the generated images are closer to the real images in terms of visual fidelity. In the context of floorplan generation, the entire test set of generated floorplans and their corresponding real counterparts are considered for FID computation. As a result, the FID score can also indirectly account for diversity, as low diversity in generated images may lead to a poor match with the real image distribution. For direct measurement of diversity in generated images, metrics such as Precision and Recall for Distributions (PRD)\cite{sajjadi2018assessing, kynkaanniemi2019improved} and Intra-FID (i.e., FID between generated images and real images within each class)~\cite{miyato2018cgans} can be used. 

 \paragraph{Compatibility} 
Compatibility with input graphs is assessed using Graph Edit Distance (GED)~\cite{abu2015exact}, a metric that evaluates whether the generated floorplan maintains the correct connectivity relationships specified by the input graph.
GED measures the total number of connectivity errors in the placement of interior doors within rooms, or front doors with respect to rooms and outdoor areas, in a generated floorplan.
\paragraph{Boundary Intersection-over-Union~(IOU)} For methods that use residential boundaries as constraints, the Intersection over Union (IoU) score between the generated room boundary and the ground truth is used as a quantitative metric to measure the overlap. A higher IoU score indicates better alignment of the generated house boundary with the input boundary.

\begin{figure*}[tb]
    \centering
    \scriptsize
    \setlength{\tabcolsep}{2pt}
    \begin{tabular}{cccc}

    \centering
        {\includegraphics[width=0.17\linewidth]{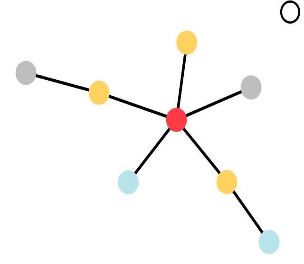}} &  
        {\includegraphics[width=0.17\linewidth]{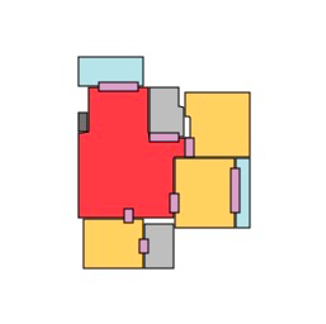}} &  
        {\includegraphics[width=0.17\linewidth]{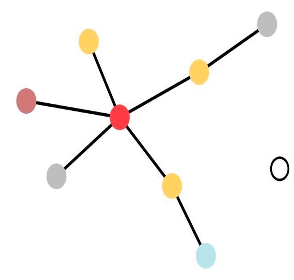}} & 
        {\includegraphics[width=0.17\linewidth]{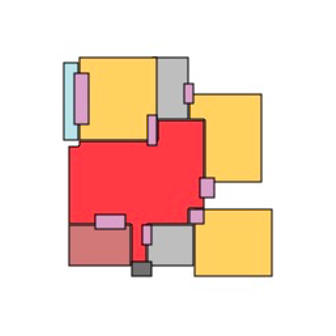}}\\

        {\includegraphics[width=0.17\linewidth]{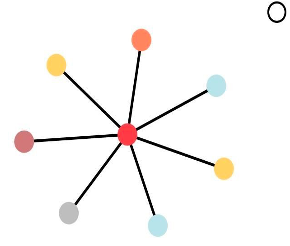}} &  
        {\includegraphics[width=0.17\linewidth]{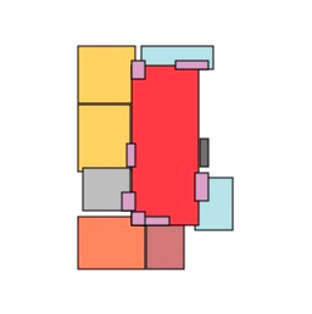}} &  
        {\includegraphics[width=0.17\linewidth]{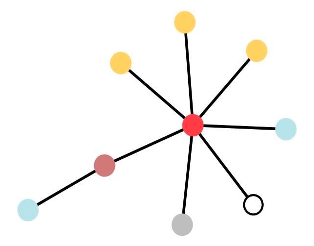}} & 
        {\includegraphics[width=0.17\linewidth]{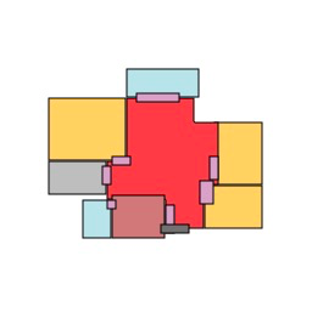}}\\

        {\includegraphics[width=0.17\linewidth]{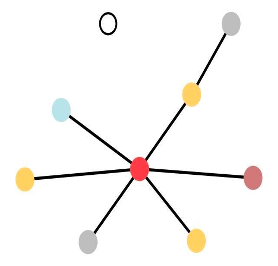}} &  
        {\includegraphics[width=0.17\linewidth]{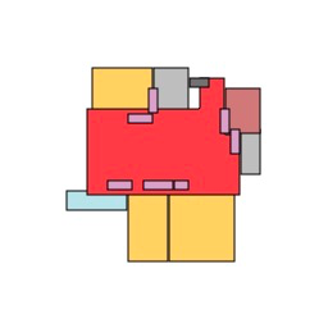}} &  
        {\includegraphics[width=0.17\linewidth]{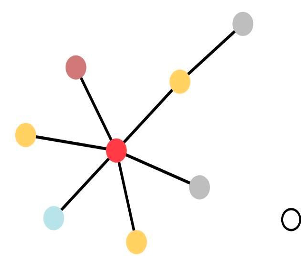}} & 
        {\includegraphics[width=0.17\linewidth]{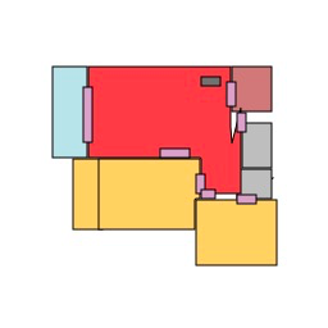}}\\
        
        {\includegraphics[width=0.17\linewidth]{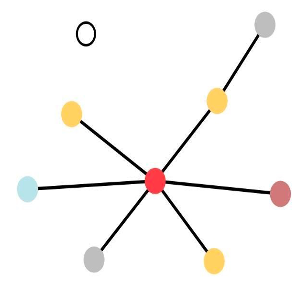}} &  
        {\includegraphics[width=0.17\linewidth]{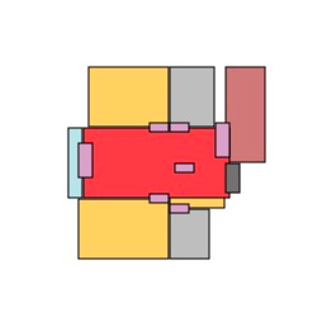}} &  
        {\includegraphics[width=0.17\linewidth]{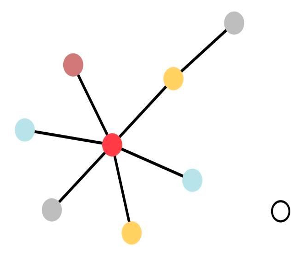}} & 
        {\includegraphics[width=0.17\linewidth]{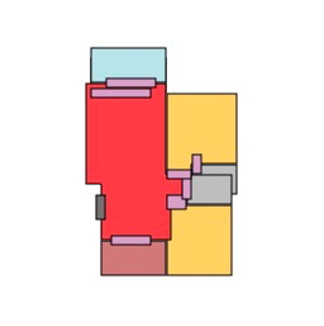}}\\

        Input Bubble Diagram & Generated Floorplan & Input Bubble Diagram & Generated Floorplan \\

    \end{tabular}
    \caption{The generated floorplans from a state-of-the-art generative model HouseDiffusion~\cite{shabani2022housediffusion}. We got the results from the most challenging setting on 8-room generation during inference, while the model is trained on the remaining data splits with other room numbers. The top two rows show satisfactory outcomes whereas the last two rows exist noticeable issues.}
    \label{fig::housediffusion_qualitative}
\end{figure*}

\subsection{Discussions}

We provide an overview of the existing methods, floorplan representations, frameworks used, supported user inputs, benchmark datasets, and the publication venues for learning-based floorplan generation methods in Tab.~\ref{tab:floorplan_methods}.

\par
For the floorplan generation methods conditioned on bubble diagrams, we list the quantitative outcomes for current SOTA learning-based methods in Tab.~\ref{tab:quantitative_floorplan}. One can see that, GAN-based methods using the rasterized representation fall behind in terms of both diversity and compatibility compared with the vectorized approaches. The multi-stage GNN and transformer framework~\cite{liu2022end} strikes a better balance between generative fidelity and computational cost. 
HouseDiffusion~\cite{shabani2023housediffusion} achieves SOTA generation quality by leveraging iterative optimization via a well-designed denoising process. To further demonstrate the generation fidelity of HouseDiffusion, we run their model using their released code\footnote{HouseDiffusion - \href{https://github.com/aminshabani/house_diffusion}{https://github.com/aminshabani/house\_diffusion}}, and showcase some representative generated sampled in Fig.~\ref{fig::housediffusion_qualitative}. The model can represent the rooms as complex polygons~(illustrated by the living rooms) and most of the rooms preserve spatial relationships consistent with the specification in the bubble diagram. For the samples of the top $2$ rows, the model has demonstrated the ability to handle the complex input bubble diagrams and succeed in generating visually appealing floorplans. However, the samples in the last $2$ rows display noticeable issues, including artifacts in spatial arrangement. The interior doors are sometimes incorrectly arranged inside the rooms~(the left sample in the 3rd and 4th rows). The generated rooms occasionally exhibit unusual wall boundaries that do not appear in real floorplans~(the right sample in the 3rd row). 
Moreover, the overall spatial arrangement of the rooms is not always optimal, with occasional unnecessary overlap between rooms~(the right sample in the 4th row).
\par
As demonstrated above, there exist non-negligible issues or limitations for current SOTA deep generative models. First, the incorrect arrangement of elements and the presence of generated artifacts should be addressed by enhancing the model’s generative capacity. Additionally, the model should better capture correct spatial and geometric inter-room relationships during generation. Second, the supported user input formats are limited. In addition to boundaries and bubble diagrams, common options for user interaction and customization include text, room sketches, and audio instructions. A generation framework capable of incorporating diverse input modalities would offer greater flexibility and be more suitable for industry-level products and applications.

\section{Scene Layout Generation}
\label{scene_synthesis}

\begin{table*}[tbp]
\caption{Indoor scene synthesis methods categorized by the type of user input, designed framework and utilized benchmark datasets and publishing press. }
\resizebox{1.0\linewidth}{!}{
\begin{tabular}{l|c|c|c|c}
\hline
{Methods} & {Framework} & {User Input} & {Benchmark Datasets} & {Publishing Press}\\
\hline
Wang~\etal & CNNs & Input Partial Scene & SUNCG & TOG 2018\\
PlanIT & Graph-based CNNs & Room Boundary & SUNCG & TOG 2019\\
GRAINS & Graph-based VAE & Unconditioned & SUNCG & TOG 2019\\
SceneGraphNet & Message Passing Networks, GRU & Unconditioned & SUNCG & ICCV 2019\\
Ritchie~\etal & CNNs & Room Boundary & SUNCG & CVPR 2019\\
3D-SLN & Graph-based VAE & Scene Graph & SUNCG & CVPR 2020\\
SG-VAE & Grammar VAE & Unconditioned & SUNCG & ECCV 2020\\
Zhang~\etal & Feed-forward Network & Unconditioned & SUNCG & TOG 2020\\
SceneHGN & Hierarchical graph-based VAE & Room Boundary & 3D-FRONT & TPAMI 2021\\
ATISS & Transformers & Room Boundary & 3D-FRONT & NeurIPS 2021\\
Depth-GAN & GANs & Unconditioned & Structured3D, Matterport3D & ICCV 2021\\
Yang~\etal & Bayesian Networks & Scene Graphs & 3D-FRONT, SUNCG & ICCV 2021\\
Graph-to-3D & Graph-based VAE & Scene Graphs & 3DSSG & ICCV 2021\\
Sceneformer & Transformers & Room Boundary & SUNCG & 3DV 2021\\
LayoutEnhancer & Transformers  & Layout Subsets & 3D-FRONT & SIGGRAPH Asia 2022\\
DiffuScene & Diffusion Models & Room Boundary & 3D-FRONT & Arxiv 2023\\
COFS & Transformers  & Layout Subsets & 3D-FRONT & SIGGRAPH 2023\\
LEGO-Net & Transformers & Messy State Layout & 3D FRONT & CVPR 2023\\
CC3D & Neural Radiance Fields, StyleGAN & 2D Floorplan & 3D FRONT, KITTI 360 & ICCV 2023\\
LayoutGPT  & LLMs  & Texts & 3D-FRONT & NeurIPS 2023\\
CommonScenes & GCNs, Diffusion Models & Scene Graphs, Texts & SG-FRONT~(3D-FRONT) & NeurIPS 2023\\
PhyScene & Diffusion Models & Room boundary & 3D-FRONT, 3D-FUTURE & CVPR 2024 \\
LTS3D & Diffusion Models & Unconditioned & 3D-FRONT & Arxiv 2024 \\
Forest2Seq & Transformers & Room boundary & 3D-FRONT & ECCV 2024\\
\hline
\end{tabular}
}
\centering
\label{tab:scene_synthesis}
\end{table*}

\begin{figure*}[!htb]
    \centering
    \includegraphics[width=0.9\linewidth]{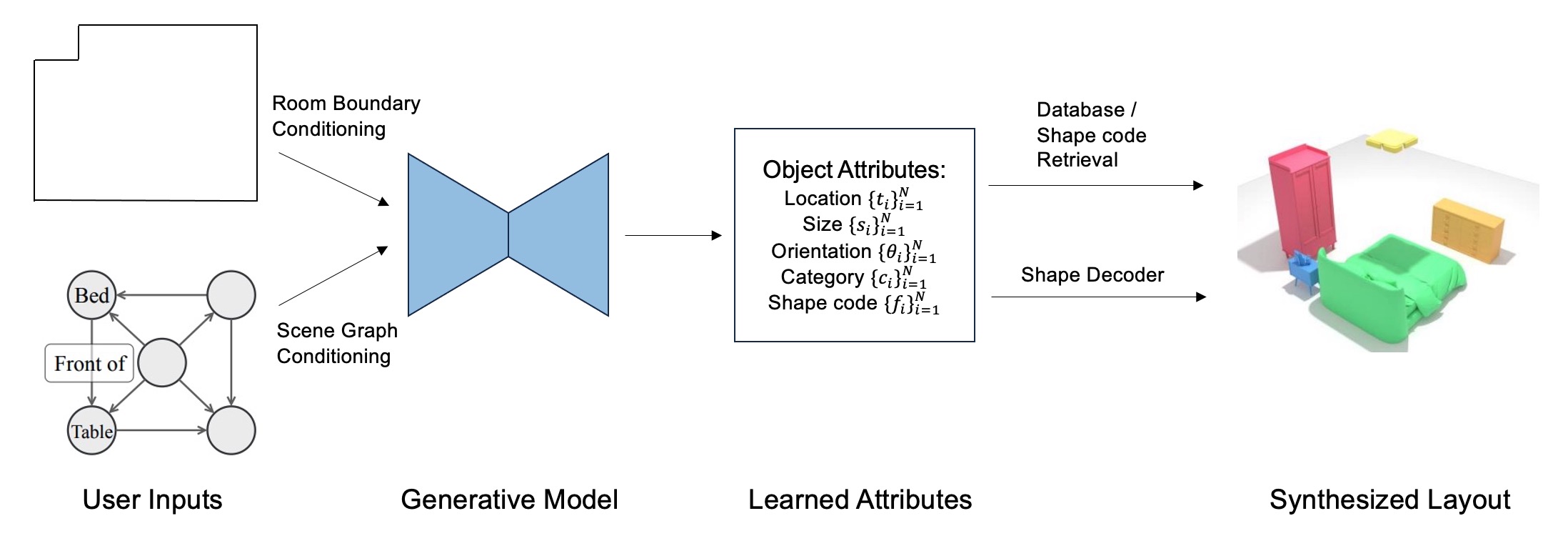}
    \caption{A generic scene synthesis framework with different types of user input. The generative model takes a room boundary or a scene graph as input condition, then generates a group of parameterized furniture object attributes. For retrieval based methods, the learned shape attributes are used for a shape retrieval stage to get object meshes. For end-to-end approaches, the shape meshes are parsed from a learned shape decoder. The icons of furniture object layouts are referred from DiffuScene.}
    \label{fig::scene_synthesis_pipeline}
\end{figure*}

We take a step further into the interior layout generation problem after establishing the floorplan wall boundaries. The placement of diverse furniture categories based on the synthesized floorplans is another crucial aspect of indoor layout design. In this section, we introduce and formulate the task of indoor scene synthesis. We further categorize data-driven methods into graph-based approaches and sequential generation techniques. Finally, we present the publicly available benchmark datasets used in this field, common evaluation metrics for this task, and discussion and quantitative comparison of scene layout generation methods. A generic methodological framework is shown in Fig.~\ref{fig::scene_synthesis_pipeline}.

\subsection{Task formulation}

Conditioned by some given user input constraints, \ie, the room boundary $F$ or scene graph $G$, the room category~(e.g. bedroom, living room)~$C$, and sometimes some input text instructions $T$ or given incomplete object layouts $L_{p}$, scene synthesis aims to generate or complete the entire scene layout sequence~$L=\{O_{i}\}_{i=1}^{N}$, ensuring correct furniture arrangement and functionality for daily activities. More specifically, a furniture object $O_{i}$ is parameterized as a 3D bounding box with parameterized location $t_{i}$, orientation $r_{i}$, dimensional size $s_{i}$, categorical type $c_{i}$ and a shape latent code $f_{i}$. Denote the designed generative model as $M$, the generation process can be formulated as:
\begin{equation}
\begin{aligned}\label{eqn:seq2}
L = \{O_{i}\}_{i=1}^{N} =\{t_{i}, r_{i}, s_{i}, c_{i}, f_{i}\}_{i=1}^{N} = M(F, G, C, T, L_{p}).
\end{aligned}
\end{equation}
Then, a shape retrieval process is typically employed to obtain the textured object mesh from the database that best matches the generated attributes, integrating them into the final scene layout.

\subsection{Methods}
\subsubsection{Optimization-based Methods}
Traditional scene modeling and synthesis works usually address this problem as an optimization process. Merrell~\etal~\cite{merrell2011interactive} identifies a set of design guidelines for house layout design and represents the scene distribution as a density function, then employs the Markov chain Monte Carlo sampler~(MCMC) to get optimized outcome while respecting user input as layout constraints. Yeh~\etal~\cite{yeh2012synthesizing} formulates the open-world layout synthesize problem as a open universe probability distribution sampling process constrained by factor graphs and proposes a reversible jump MCMC method to represent and and solve the optimization problem. Yu~\etal~\cite{yu2011make} first extracts spatial and hierarchical relationship of the objects and obtains an plausible initialization. Then the furniture arrangements are refined through a simulated annealing optimization stage. Moreover, Zhang~\etal~\cite{zhang2023automatic} proposes a system that iteratively adds patterns with respect to constraints formulated by commercial design rules by employing optimization procedures. A group of works first parse structural scene graphs from natural languages~\cite{fisher2010context, chang2017sceneseer, chang2014learning} or sketches~\cite{xu2013sketch2scene} as constraints, then synthesize the scene layouts by querying and matching the database with the learned spatial knowledge. Zhang~\etal~\cite{zhang2021fast} optimizes the layout with learned room geometry and object distribution priors, partitioning the input objects into disjoint groups, followed by layout optimization using position-based dynamics (PBD) based on the Hausdorff metric. Some works~\cite{fu2017adaptive, fisher2015activity, jiang2012learning} manage to model the distribution of human activities and incorporate such contexts into the scene layout generation process. Although having shown desirable fidelity under certain scenarios, these methods are driven by the prior knowledge of the scene distribution and regularized by hand-crafted guidelines. As a result, generation diversity is generally inferior to that of recent deep generative model-based methods, which will be discussed in the next subsection.

\begin{figure}[!htb]
    \centering
    \includegraphics[width=1.0\linewidth]{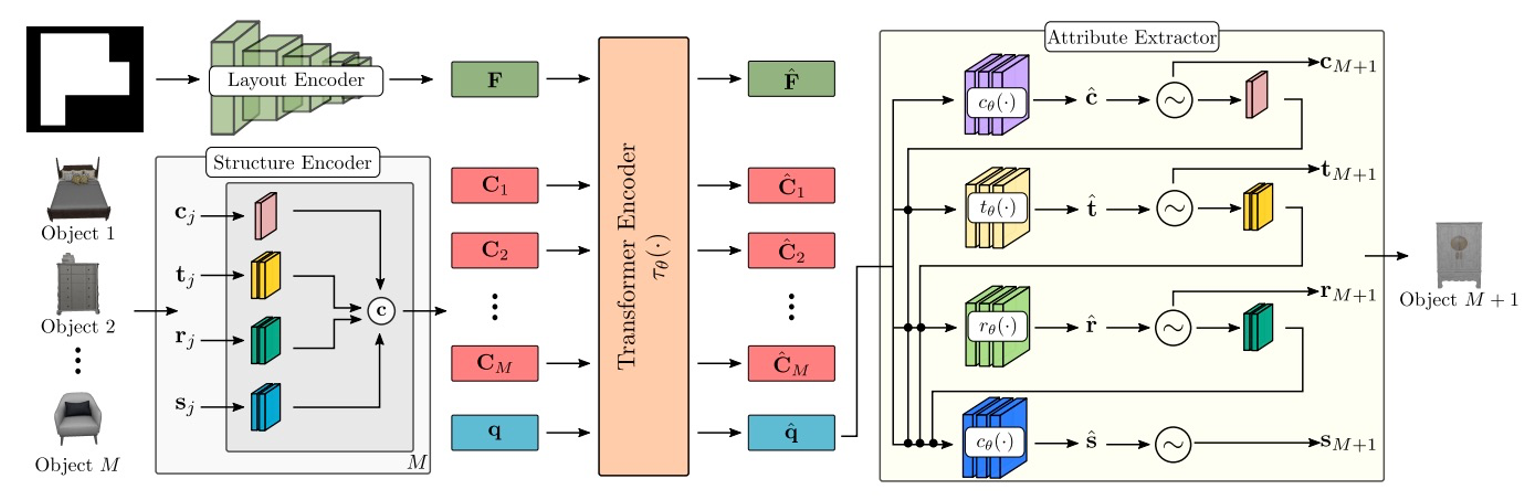}
    \caption{ATISS serves as a representative indoor scene synthesis pipeline. The top-down view residential boundary is passed into a layout encoder to extract boundary feature $F$. Different attributes of the furniture objects $c$, $t$, $r$ and $s$ are encoded by a structure encoder to learn respective embeddings regardless of the object sequence order. Then a transformer encoder processes these learned embeddings and output the refined embeddings. Then separate attribute extractors are designed for decoding the distributions of these attributes. During inference, the object attributes are sampled from the learned distributions.}
    \label{fig::atiss}
\end{figure}

\subsubsection{Learning-based Methods}
We categorize the methods of data-driven scene layout synthesis into feed-forward approaches and autoregressive-model-based approaches. Feed-forward methods are usually designed as graph-based networks, treating the furniture objects together with the floor boundary as a scene graph or other structural representation to exploit their contextual relationship, or seek to learn a powerful scene representation with a simple yet effective feed-forward generative network. Autoregressive-based methods regard the target objects as an ordered or unordered sequence, and generate the objects in an iterative manner, taking the previous generation as the context of the next step. 

\paragraph{Graph-based methods}
The objects in a scene are not independent but interleave with each other, indicating rich spatial relationships and context information. To this end, graph-based methods~\cite{wang2019planit, luo2020end, li2019grains, purkait2020sg, zhang2020deep, zhou2019scenegraphnet, yang2021scene} are proposed to solve the scene synthesis problems, which aim to first represent the scene layout as an abstracted graph, then exploit scene contexts and inter-object relationships via the built graph. The generated layout are naturally conformed to the relationships suggested in the scene graph. PlanIT~\cite{wang2019planit} proposes to first extract a scene graph trained by a deep generative model, then employs another image-based reasoning model to iteratively insert objects into the scene. GRAINS~\cite{li2019grains} formulates the scene as a hierarchical graph, then employs a recursive VAE on learning object group encoding and generation decoding. 3D-SLN~\etal~\cite{luo2020end} proposes a variational generative model to synthesize layout from scene graphs based on GCNs and VAE. SG-VAE~\cite{purkait2020sg} represents the scene as a tree structure and leverages grammar-based auto-encoder to learn the co-occurrence and appearance attributes of the scene. Depth-GAN~\cite{yang2021indoor} introduces a GAN-based framework to learn a 3D scene representation for scene synthesis through projection supervision from a set of 2.5D semantic-segmented depth images. Zhang~\etal~\cite{zhang2020deep} proposes a hybrid representaion which capture information from both 3D object and image-based representation with a feed-forward generative model. SceneGraphNet~\cite{zhou2019scenegraphnet} designs a deep and dense message passing network on extracting structural and spatial relationships, to obtain a probability distribution among object categories given a query location. Yang~\etal~\cite{yang2021scene} proposes a Bayesian optimization framework which first generates over-complete sets of attributes then employs a pruning stage to filter out the infeasible predictions based on consistency constraints of the attributes. SceneHGN~\cite{gao2023scenehgn} employs a recursive VAE network to learn the scene layout hierarchy from room to object and partial object level. Most recently, DiffuScene~\cite{tang2024diffuscene} first builds a complete 3D scene graph to encode the spatial context then represents and generates the object attributes with a diffusion network~\cite{ho2020denoising}. Different from prior work which generates object attributes then apply a database retrieval process to obtain textured object mesh which heavily restricts the generation capacity, Graph-to-3D~\cite{dhamo2021graph} proposes a GCN-based VAE architecture to learn to directly generate object 3D meshes given the scene graph as input with an end-to-end framework. To better handle the biased generation results due to the object category imbalance in the training set, FairScene~\cite{wu2024fairscene} exploits unbiased object interactions with a causal reasoning framework which achieves fair scene synthesis by calibrating the long-tailed category distribution.

\begin{figure}[!htb]
    \centering
    \includegraphics[width=1.0\linewidth]{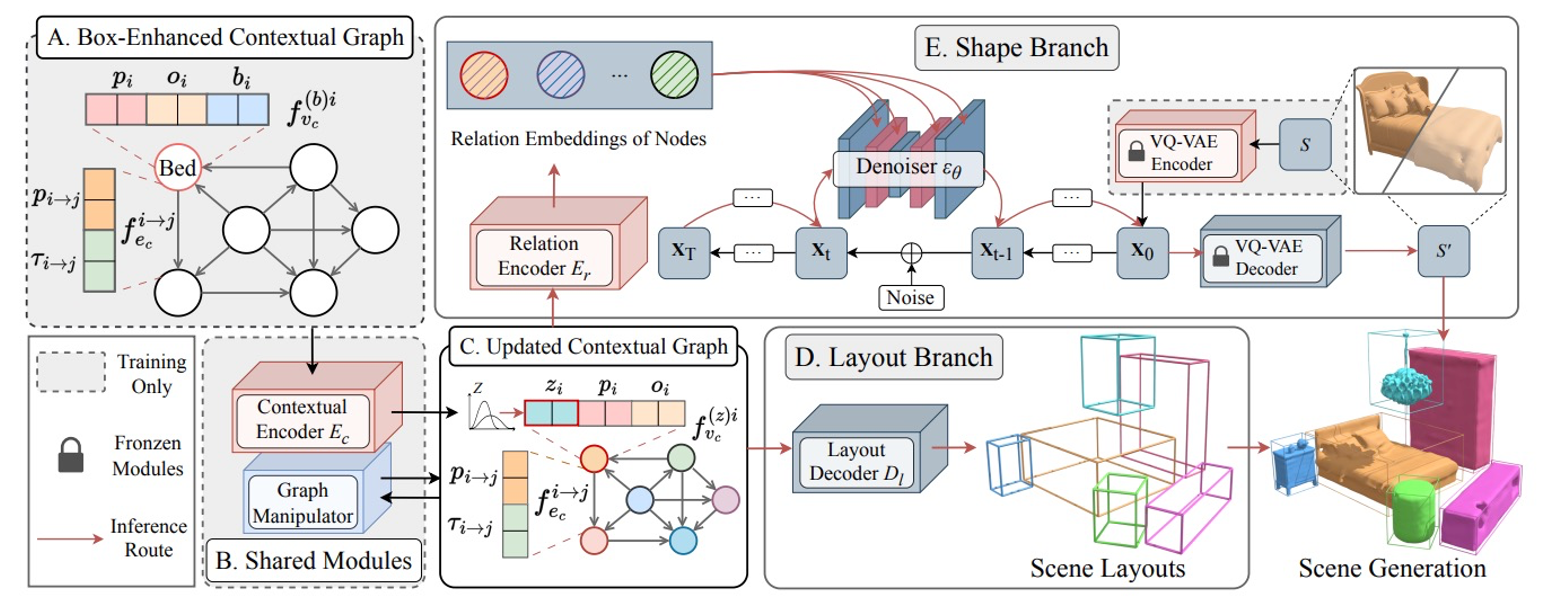}
    \caption{CommonScenes is one of the SOTA indoor scene layout generation pipelines. It leverages the CLIP-enhanced GCN feature encoder to process the scene graph constraints, a GCN decoder to synthesize layouts, and a diffusion model to decode the object mesh. In this method, the furniture meshes are generated with an end-to-end framework without any database or shape code retrieval stage.}
    \label{fig::commenscenes}
\end{figure}

\paragraph{Sequential generation methods}
Graph-based methods need to build scene graphs based on prior knowledge as prerequisites. Contrastively, auto-regressive models do not require such prior knowledge, and directly treat the objects as an ordered or unordered sequence, generating the target furniture in an iterative manner. Wang~\etal~\cite{wang2018deep} proposes to first parse the top-down view input partial scene and get a scene representation. Then separate CNNs are utilized to predict object attributes and determine whether to continue to insert objects, to complete the scene layout generation. Ritchie~\etal~\cite{ritchie2019fast} encodes the top-down view of the scene with a CNN and predicts the attributes of the objects in a sequential manner. Sceneformer~\cite{wang2021sceneformer} introduces a set of transformers~\cite{vaswani2017attention} to learn different targets separately and autoregressively insert objects to the scene in a pre-defined order. ATISS~\cite{paschalidou2021atiss}~(shown in Fig.~\ref{fig::atiss}) takes the house boundary layout as input constraint, and employs a transformer architecture as attribute encoder as well as separate attribute decoders to learn order-invariant sequential generation over furniture objects. In addition to auto-regressive generation, a set of work~\cite{leimer2022layoutenhancer, para2023cofs} regard the objects as a sequence but synthesize all of the objects at a time via feed-forward networks. COFS~\cite{para2023cofs} devises an order-invariant autoregressive transformer to perform cross-attention over the entire conditioning input, such that the generation can be conditioned with fine-grained input. LayoutEnhancer~\cite{leimer2022layoutenhancer} accounts for ergonomic qualities as expert knowledge into the generation process to tackle the challenges induced by imperfect training data. RoomDesigner~\cite{zhao2024roomdesigner} proposes to leverage anchor latents to encode the piece-wise geometric representation of furniture, then conduct scene generation sequentially with a transformer network. CommonScenes~\cite{zhai2024commonscenes}~(Fig.~\ref{fig::commenscenes}) constructs a new dataset dubbed SG-FRONT on top of 3D-FRONT providing conditional scene graphs, and proposes to encapsure inter-object relationship and local shape cues with a diffusion model, and employs a diffusion model to generate the 3D scene layout with diverse shapes in an end-to-end manner without any database retrieval stage or category-level decoders. Forest2Seq~\cite{sun2025forest2seq} derives ordering information from the layout sets and designs a transformer to generate realistic 3D scenes in an autoregressive manner. GLTScene~\cite{li2024gltscene} incorporates the interior design principles with learning techniques and adopts a global-to-local strategy for this task, by designing two transformer networks to learn the global and local priors, respectively. 

\paragraph{Other methods} Except for the above two mainstream paradigms, some methods build their framework upon some recent innovations such as neural radiance field~\cite{mildenhall2021nerf} to achieve 3D-aware generation, or leverage the generation ability of LLMs~\cite{ouyang2022training} as a prior. CC3D~\cite{bahmani2023cc3d} takes the 2D furnitured floorplan as the input constraint and aims to generate 3D scene layout. It proposes to leverage a neural radiance field to lift the 2D features into 3D, and use the StyleGAN~\cite{karras2019style} architecture as the generator and discriminator to synthesize the layouts. LEGO-Net~\cite{wei2023lego}, inspired by the workflow of diffusion models, aims to refine a messy scene layout into a cleaner, more organized one. It devises a denoising transformer architecture to gradually transform the original messy state to a clean state. Most recently, leveraging the rise of large language models~(LLMs), LayoutGPT~\cite{feng2024layoutgpt} proposes a LLM-based training-free layout generation framework. It first prepares well-constructed text prompts such as Cascading Style Sheets~(CSS) formats and specified task instructions, then generates desirable image or scene layouts through LLMs like GPTs~\cite{ouyang2022training}, and has shown comparable or more superior generative fidelity compared with the existing SOTAs trained on a particular dataset. Other than these, Physcene~\cite{yang2024physcene} proposes a diffusion-based framework to generate indoor scene layouts imposing constraints such as object collision, room layout, and object reachability, integrating embodied AI environment into the scene synthesis task. LT3SD~\cite{meng2024lt3sd} introduces a latent tree representation coupling the diffusion models to generate complex 3D scene geometry. FuncScene~\cite{min2024funcscene} proposes a VAE network which leverages function groups as an intermediate representation to connect the local scenes and the global structure, aiming to achieve a coarse-to-fine indoor scene synthesis.

\begin{table*}[!htb]
\caption{Quantitative comparison on retrieval-based deep scene layout generative methods. The numerical results are referred from the latest method DiffuScene. }
\centering
\resizebox{1.0\linewidth}{!}{
\setlength{\tabcolsep}{4pt}{
\begin{tabular}{l|cccc|cccc|cccc}
\hline
\multirow{2}{*}{Methods} & \multicolumn{4}{c|}{Bedroom} & \multicolumn{4}{c|}{Dining Room} & \multicolumn{4}{c}{Living Room}\\
& FID~($\downarrow$) & KID~($\downarrow$) & SCA~($\uparrow$) & CKL~($\downarrow$) & FID~($\downarrow$) & KID~($\downarrow$) & SCA~($\uparrow$) & CKL~($\downarrow$) & FID~($\downarrow$) & KID~($\downarrow$) & SCA~($\uparrow$) & CKL~($\downarrow$)\\
\hline 
DepthGAN & 40.15 & 18.54 & 96.04 & 5.04 & 81.13 & 50.63 & 98.59 & 9.72 & 88.10 & 63.81 & 97.85 & 7.95\\
Sync2Gen & 31.07 & 11.21 & 82.97 & 2.24 & 46.05 & 8.74 & 88.02 & 4.96 & 48.45 & 12.31 & 84.57 & 7.52\\
Sync2Gen$^{*}$ & 33.59 & 13.78 & 87.11 & 2.67 & 48.79 & 12.01 & 91.43 & 5.03 & 47.14 & 11.42 & 86.71 & 1.60 \\
ATISS & 18.60 & 1.72 & 61.71 & 0.78 & 38.66 & 5.62 & 71.34 & 0.64 & 40.83 & 5.18 & 72.66 & 0.69\\
DiffuScene & 17.21 & 0.70 & 52.15 & 0.35 & 32.60 & 0.72 & 55.50 & 0.22 & 36.18 & 0.88 & 57.81 & 0.21\\
\hline
\end{tabular}
}}
\label{tab:quantitative_scene_synthesis_retrieval_1}
\end{table*}

\begin{figure}[!htb]
    \centering
    \includegraphics[width=1.0\linewidth]{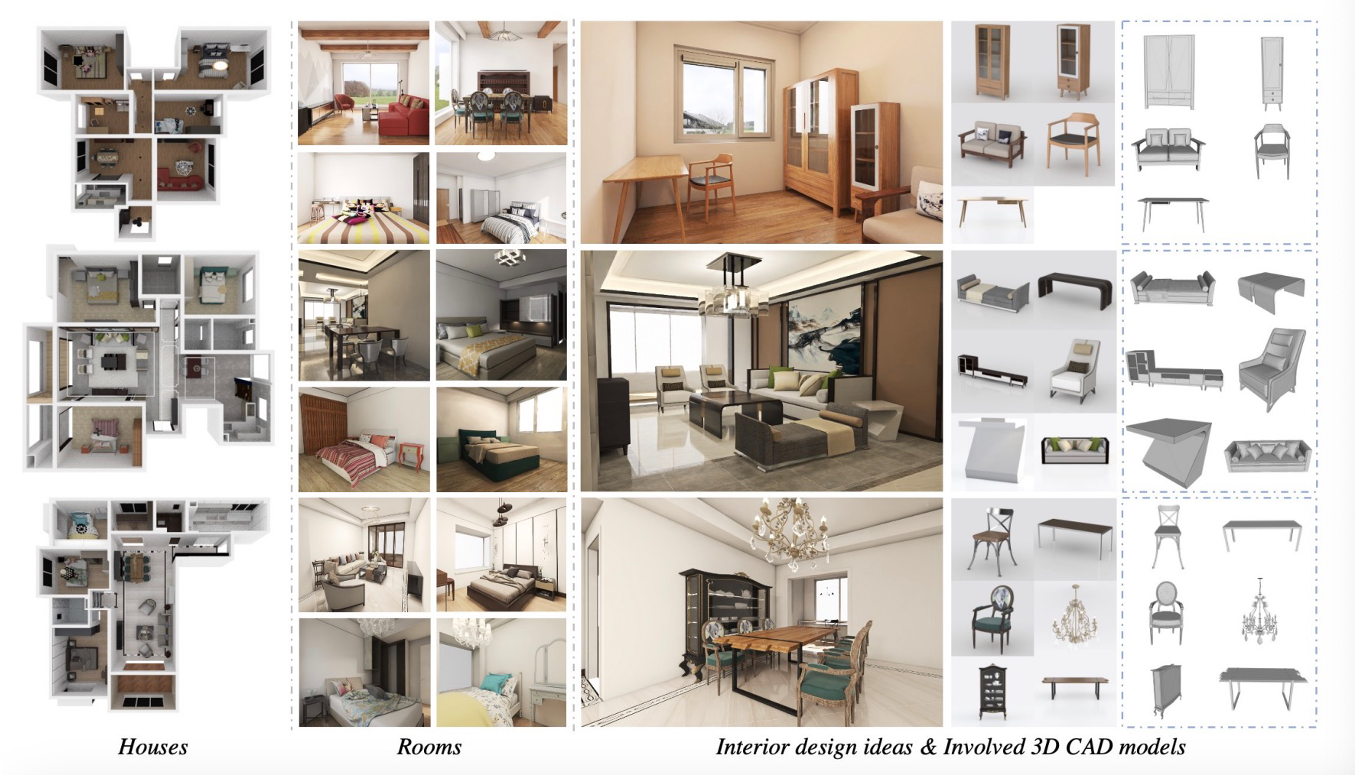}
    \caption{3D-FRONT provides a large-scale interior house layout database with versatile house and room layouts, as well as intricate furniture CAD models.}
    \label{fig::3d_front}
\end{figure}

\begin{table*}[!htb]
\caption{Quantitative comparison on scene synthesis methods which take scene graphs as input. The numerical results are referred from a latest method CommonScenes.}
\centering
\resizebox{1.0\linewidth}{!}{
\setlength{\tabcolsep}{4pt}{
\begin{tabular}{l|l|cc|cc|cc|cc}
\hline
\multirow{2}{*}{Methods} & \multirow{2}{*}{Shape Representation} & \multicolumn{2}{c|}{Bedroom} & \multicolumn{2}{c|}{Living Room} & \multicolumn{2}{c|}{Dining Room} & \multicolumn{2}{c}{All}\\
& & FID~($\downarrow$) & KID~($\downarrow$) & FID~($\downarrow$) & KID~($\downarrow$) & FID~($\downarrow$) & KID~($\downarrow$) & FID~($\downarrow$) & KID~($\downarrow$)\\
\hline
3D-SLN & Retrieval & 57.90 & 3.85 & 77.82 & 3.65 & 69.13 & 6.23 & 44.77 & 3.32 \\
Progressive & Retrieval & 58.01 & 7.36 & 79.84 & 4.24 & 71.35 & 6.21 & 46.36 & 4.57 \\
Graph-to-Box & Retrieval & 54.61 & 2.93 & 78.53 & 3.32 & 67.80 & 6.30 & 43.51 & 3.07 \\
\hline
Graph-to-3D & DeepSDF & 63.72 & 17.02 & 82.96 & 11.07 & 72.51 & 12.74 & 50.29 & 7.96 \\
Layout+txt2shape & SDFusion & 68.08 & 18.64 & 85.38 & 10.04 & 64.02 & 5.08 & 50.58 & 8.33 \\
CommonScenes & rel2shape & 57.68 & 6.59 & 80.99 & 6.39 & 65.71 & 5.47 & 45.70 & 3.84 \\
\hline
\end{tabular}
}}
\label{tab:quantitative_scene_synthesis_retrieval_2}
\end{table*}

\subsection{Benchmark Datasets}
\paragraph{3D-FRONT}
3D-FRONT~\cite{fu20213d}~(as shown in Fig.~\ref{fig::3d_front}) is a large-scale indoor scene and object layout dataset containing diverse room categories and high-quality textured 3D models with different styles. The layout designs are sourced from professional creations while the furniture texture and styles are managed by a recommend system to ensure consistent and expert designs. This dataset is widely adopted as a benchmark for indoor scene or texture synthesis applications.
\paragraph{3D-FUTURE}
3D-FUTURE~\cite{fu20213dfuture} complements 3D-FRONT by providing high-quality 3D shapes, informative textures and attributes. Additionally, 3D-FRONT fill the blank of the large-scale and accurate 2D-3D alignment between realistic image and 3D objects by rendering over 20K photo-realistic images across diverse scenes. This enables its promising applications on serving as benchmarks onto tasks such as high-quality 3D shape generation, reconstruction and retrieval. 
\paragraph{3DSSG} 3DSSG~\cite{wald2020learning} is a large scale 3D dataset built on top of 3RScan~\cite{wald2019rio} with semantic scene graph annotations, containing spatial relations between objects, classes and other attributes. In total it contains 363k graph-image pairs, which are rendered from 3D scene graphs. This dataset enables large-scale training on scene layout synthesis conditioned on scene graphs.
\paragraph{SG-FRONT}
SG-FRONT~\cite{zhai2024commonscenes} is a dataset created in CommonScenes~\cite{zhai2024commonscenes} and is developed from 3D-FRONT~\cite{fu20213d} dataset. It aims to encourage future exploration on conditional scene synthesis on scene graphs. To this end, it offers a set of well-annotated scene graph labels grouped into three categories: spatial/proximity, support, and style. SG-FRONT covers 15 relationship types densely annotating scenes. More details can be found in the original paper.

\subsection{Evaluation Metrics}
As shown in Tab.~\ref{tab:quantitative_scene_synthesis_retrieval_1} and Tab.~\ref{tab:quantitative_scene_synthesis_retrieval_2}, the commonly-used evaluation metrics in mainstream approaches on leading benchmarks include FID, KID, SCA and CKL, assessing fidelity, diversity, and adherence to graph constraints. 

\paragraph{Fidelity and diversity.} Generation fidelity and diversity are common standards for image synthesis tasks. The generated 3D scene layout is first projected as images from multiple bird-eye views, then the generation quality is measured by the Fréchet Inception Distance~(FID)~\cite{heusel2017gans} and Kernel Inception Distance~\cite{binkowski2018demystifying} between the synthesized layout and the real ones. Besides, the KL divergence~\cite{akaike1998information} between the object category distributions of the generated and groundtruth furniture serves as another quantitative metric. Moreover, scene classification accuracy~(SCA) is used to measure whether the synthesized layouts is good enough to be indistinguishable from real scenes.
\paragraph{Graph constraint metrics.} For the methods which take scene graphs as input constraints, object pair-wise accuracy on the assigned constraints are evaluated to achieve consistency with the input constraints. The constraints contain three categories: spatial/proximity~(\eg, left/right, bigger/smaller), support~(\eg, close by, above), and semantic or style level relationships~(\eg, same material as, same category as), etc. More details can be referred to from~\cite{zhai2024commonscenes}. 
\subsection{Discussions}
We list the learning-based scene layout synthesis methods, their frameworks, enabled user inputs, benchmark datasets, and publishing venues in Tab.~\ref{tab:scene_synthesis}. It can be observed that autoregressive transformers, which represent the scene as a sequence, and graph-based generators, which represent the scene as a graph, are two mainstream frameworks for learning object attributes. More recently, researchers have also explored the potential of large language models (LLMs) and diffusion models to facilitate scene layout generation.
A key limitation of current methods is scalability. Most focus solely on single-room layout generation, rather than house-level generation involving multiple rooms. The relationships between rooms may provide additional spatial and semantic cues for object arrangement. Another limitation is that most methods involve a separate retrieval stage to obtain object meshes from the database after learning object attributes. While this ensures plausible object shapes, it significantly limits generation diversity. Although some works~\cite{dhamo2021graph, zhai2024commonscenes} propose generating object layouts and meshes in an end-to-end manner, the resulting meshes still exhibit noticeable artifacts or unavoidable collisions between objects. Optimizing layout attributes and decoding high-fidelity shapes more effectively in a fully end-to-end pipeline remains an open area for further exploration.

\section{Other Building Layout Design and Synthesis}
\label{other}
Sections~\ref{floorplan} and~\ref{scene_synthesis} primarily focus on the generation and synthesis processes of apartment-level (or flat-level) layouts. In contrast, this section provides an overview of layout generation in a broader architectural context, extending from individual buildings to site-level considerations such as blocks or parcels in communities. Due to the distinct objectives associated with these tasks, there is a lack of shared or unified benchmark datasets and evaluation protocols. Consequently, this section does not delve into the specifics of datasets and evaluations. Interested readers are encouraged to consult relevant works for more detailed information.

\subsection{Building-level layout generation}

We begin by expanding the task scope from individual apartments to entire buildings. This expansion encompasses the comprehensive planning of multiple apartments, staircases, elevators, and other public amenities across entire floors. Simultaneously, we must address the holistic arrangement of elements within multi-story buildings. This extended scope introduces unique challenges, such as generating roofs~\cite{ren2021intuitive,qian2021roof}, shear walls~\cite{liao2021automated,lu2022intelligent,zhao2022intelligent}, facades~\cite{gerber2017multi}, and volumetric representation~\cite{chang2021building,alam2023representation}.

To address this complex domain, we synthesize relevant literature, categorizing it into non-learning-based and learning-based methodologies. This approach allows us to gather information and methods from diverse sources, leading to a deeper understanding of the complexities of designing entire building structures.

\begin{figure}[!htb]
    \centering
    \includegraphics[width=1.0\linewidth]{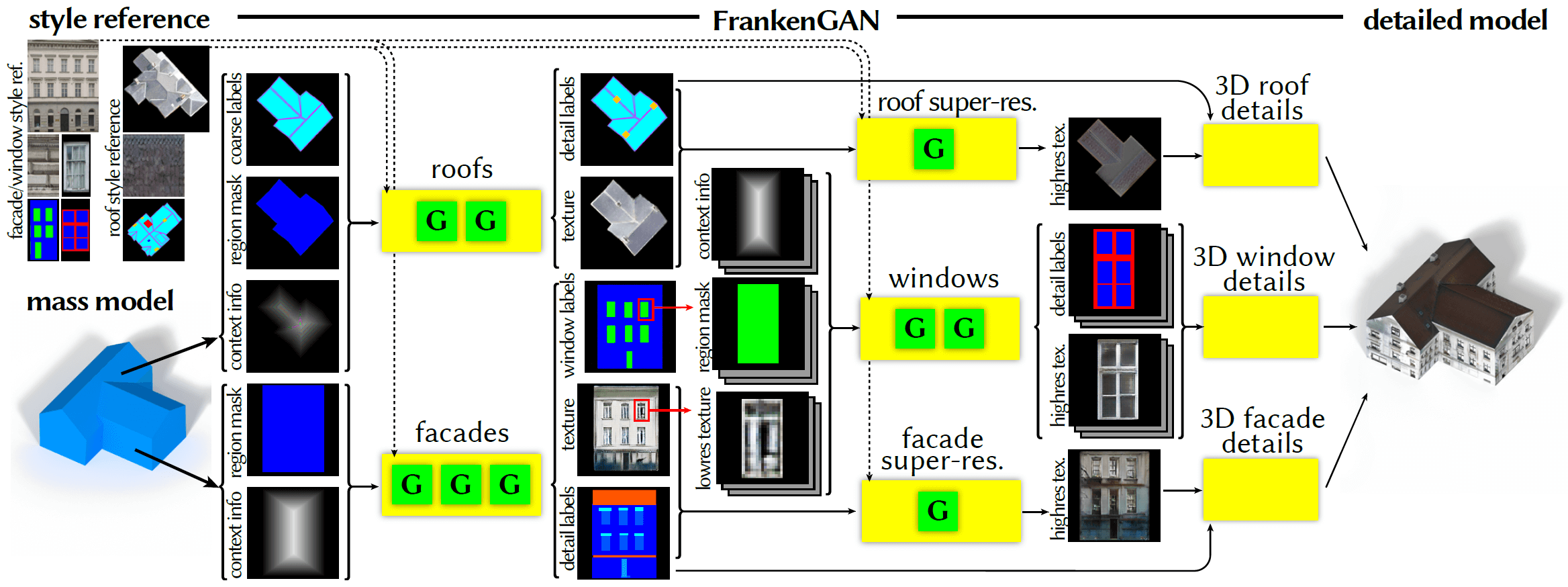}
    \caption{FrankenGAN is a framework for generating building layouts using GANs. It involves three steps: generating texture and label maps for facades and roofs, enhancing texture resolution with dedicated chains, and creating 3D details for roofs, windows, and facades based on the generated maps.}
    \label{fig:frankengan}
\end{figure}

\begin{figure*}[tb]
    \centering
    \includegraphics[width=1.0\linewidth]{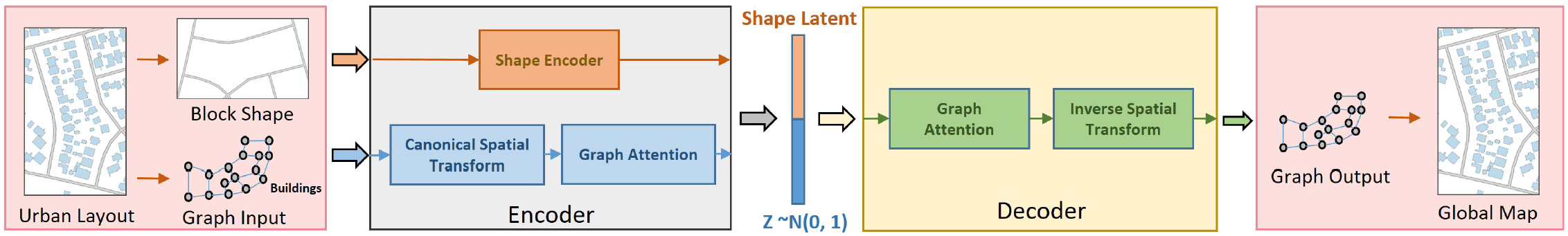}
    \caption{GlobalMapper is a representative method for site-level layout generation. It uses a VAE framework with a CNN encoder to transform arbitrary block shapes into a binary mask in latent space, allowing for the conditional generation of large urban layouts.}
    \label{fig:globalmapper}
\end{figure*}

\subsubsection{Non-learning-based methods}

M{\"u}ller~\etal~\cite{muller2006procedural}  introduced CGA shape grammar to create visually appealing and detailed building shells with consistent mass models in computer-generated architecture.
Following that, Schwarz and M{\"u}ller~\cite{schwarz2015advanced} introduced CGA++ to procedurally model architecture, providing an integrated and versatile solution that builds upon the CGA shape.
Bao~\etal~\cite{bao2013generating} first located and represented good layouts as portal graph, then used this portal graph and local shape grammars to explore more building layouts.
Rodrigues~\etal~\cite{rodrigues2013approach} introduced a way to handle different levels using a mix of evolutionary techniques, where stairs and elevators adapt and interact with other spaces in the search process.
Bahrehmand~\etal~\cite{bahrehmand2017optimizing} shared a tool that helps designers create personalized layouts based on guidelines and user preferences, with key features including addressing subjective design aspects and using a genetic algorithm for better quality layouts.
Wu~\etal~\cite{wu2018miqp} proposed a structured approach to create indoor spaces using a mixed integer quadratic programming (MIQP) method, applicable to various building sizes like homes, offices, malls, and supermarkets.
Gan~\cite{gan2022bim} introduced a graph data model using Building Information Modeling (BIM) to represent key features in modular buildings.
Isaac~\etal~\cite{isaac2016methodology} proposed a graph-based approach for offsite preassembly with clustering algorithm on BIM tool data, and providing a computer program for automated application in large and complex projects.
Sharafi~\etal~\cite{sharafi2017automated} proposed a Unified Matrix Method for efficiently finding the optimal spatial design of multi-story modular buildings in early design stages.
Fan~\etal~\cite{fan2023automated} design a two-stage genetic algorithm~(GA) to automatically generate modular high-rise residential buildings~(MHRBs).
Gerber~\etal~\cite{gerber2017multi} investigated using multi-agent systems (MAS) in architectural design to enhance and partially automate the design process.

\subsubsection{Learning-based methods}

Advanced techniques in deep generative models have made scalable, building-level layout generation possible.
Merrell~\etal~\cite{merrell2010computer} presented a Bayesian network method for automatically creating detailed multi-story building layouts based on high-level requirements.
Ghannad and Lee~\cite{ghannad2021developing} suggested a method to automatically create and set up a modular building design using GAN and building information modeling (BIM) technologies.
They later~\cite{ghannad2022automated} introduced a framework using Coupled GAN (CoGAN) for automating the generation of modular housing designs.
Wang~\etal~\cite{wang2023automated} made a dataset of building layouts and graphs, propose a mix of methods for diverse graph matching, and create a neural network with U-Net and spatial attention for effective building region segmentation.
FrankenGAN~\cite{kelly2018frankengan} addressed the challenge of realistically detailing mass models by employing latent style vectors through synchronized GANs guided by exemplar style images, with the pipeline shown in Fig.~\ref{fig:frankengan}.
Du~\etal~\cite{du20203d} proposed a 3D building fabrication method using a multi-properties GAN chain for complex architectural structures.
Building-GAN~\cite{chang2021building} utilized a volumetric representation for buildings and proposes a graph-conditioned GAN framework to generate building layouts.
Alam~\etal~\cite{alam2023representation} treated the volumetric building design as a sequential problem and creates a sequential volumetric design dataset. Latent representation learning is utilized to tackle the auto-completion and reconstruction of design sequences.
Vitruvio~\cite{tono2024vitruvio} introduced a conditional generation framework based on sketch and context for single-view building mesh reconstruction.
Fu~\etal~\cite{fu2023dual} proposed a GAN-based approach (FrameGAN) for automating the layout design of components in steel frame-brace structures.
Liao~\etal~\cite{liao2021automated} suggested using a GAN-based method to quickly and intelligently design shear walls by learning from existing design documents.
Zhao~\etal~\cite{zhao2022intelligent} introduced an approach to design layouts for reinforced concrete structures using deep neural networks, incorporating building space and element attributes as input, and generating new designs based on learned principles.
Lu~\etal~\cite{lu2022intelligent} introduced StructGAN-PHY, a physics-boosted GAN for intelligent generative design in structural engineering, leveraging a surrogate model for physical performance assessment.
Roof-GAN~\cite{qian2021roof} produced a structured roof model as a graph, assessing the primitive geometry, inter-primitive, primitive geometry on building roofs, and generates structural roof primitives based on GANs.
Ren~\etal~\cite{ren2021intuitive} proposed a roof graph representation which encodes roof topology and employs auto-regressive models to synthesis roof structures.

\subsection{Site-level layout generation}
We then expand the scope of layout generation tasks from single-building configurations to include multiple buildings. Following~\cite{wang2023automated}, we refer to this specific challenge as site-level layout generation, involving the placement of buildings within a designated block or parcel of land. While existing works predominantly target urban planning, limited research specifically addresses rural building layouts~\cite{emilien2012procedural}. 
We organize the pertinent literature into two categories: non-learning-based methods and learning-based methods. In terms of benchmark, Reco~\cite{chen2022reco} has created and maintained the first large-scale open-source dataset for residential community layout planning, covering 37,646 residential communities and 598,728 buildings across 60 cities.

\subsubsection{Non-learning-based methods}

Parish and M{\"u}ller~\etal~\cite{parish2001procedural} proposed a city modeling system using L-systems, which processes input maps to generate roads, divide land into lots, and create building geometries.
Aliaga~\etal~\cite{aliaga2008interactive} presented a system that uses real-world urban data and synthesis algorithms to interactively generate realistic urban layouts with both structural and visual components.
Vanegas~\etal~\cite{vanegas2012procedural} introduced a technique for user-guided creation of city parcels in urban modeling, involving the subdivision of city block interiors based on specified attributes and style parameters.
Lipp~\etal~\cite{lipp2011interactive} proposed a method for interactive city layout modeling, combining procedural and manual modeling.
Emilien~\etal~\cite{emilien2012procedural} proposed a method for creating village layouts on diverse terrains, utilizing a hybrid settlement/road generation process with dynamic interest maps and an anisotropic conquest process.
Peng~\etal~\cite{peng2014computing} presented a solution for covering a domain with deformable templates, ensuring no overlap by considering constraints and permissible deformations.
Yang~\etal~\cite{yang2013urban} offered a framework for generating quality street networks and parcel layouts using hierarchical domain splitting for urban planning and virtual environments.
Wang~\etal~\cite{wang2020generative} employed a digital description framework and generative grammar to examine the morphological complexity of block forms.
Nagy~\etal~\cite{nagy2018generative} demonstrated how Generative Design optimizes a residential neighborhood project, showcasing its potential in solving complex urban design challenges with conflicting stakeholder demands.
Sung and Jeong~\etal~\cite{sung2022site} presented an efficient approach to organizing multiple buildings on a site using the C\# script component in Rhino3d and Grasshopper, offering rapid, real-time results without complex computations.

\subsubsection{Learning-based methods}

Feng~\etal~\cite{feng2016crowd} introduced a random forest-based method for designing mid-scale layouts in areas like shopping malls, optimizing paths and sites based on crowd flow metrics: mobility, accessibility, and coziness.
Liang~\etal~\cite{liang2020building} employed Generative Latent Optimization (GLO) and adversarial training to develop a model for effortless generation and placement of buildings on a designated map.
Shen~\etal~\cite{shen2020machine} employed GANs to automatically generate urban design plans, predicting building details based on city conditions.
Fedorova~\cite{fedorova2021generative} explored using GANs to design urban blocks, employing a flexible model that learns from the existing city context rather than explicitly defining parameters.
BlockPlanner~\cite{xu2021blockplanner} introduced a vectorized block-plan representation and employs a graph-constrained VAE to decode different geometric attributes.
Ying~\etal~\cite{ying2023intelligent,ying2021generating} proposed an intelligent method using genetic algorithms and CNNs to optimize high-density residential building layouts considering local wind conditions.
Quan~\cite{quan2022urban} presented Urban-GAN, a user-friendly urban design system enabling individuals with minimal design expertise to select, generate, and make design decisions based on urban form cases.
Jiang~\etal~\cite{jiang2023building} presented ESGAN, a GAN-based model for automated building layout generation, incorporating a conditional vector to meet project requirements in different design scenarios.
Sun~\etal~\cite{sun2023development} explored machine learning preferences in residential site plan layouts (RSPL) using the Pix2pix model, aiming to enhance applications in residential and urban planning development.
GlobalMapper~\cite{he2023globalmapper} represented city block layouts as graphs, encoding building layouts into a shape-independent canonical representation. As shown in Fig.~\ref{fig:globalmapper}, it also enables conditional generation on realistic urban layouts for arbitrary road networks.

Recently, He and Aliaga~\cite{he2024coho} have proposed a graph-based masked autoencoder (GMAE) that uses a canonical graph representation to generate large-scale, context-sensitive urban layouts with high realism and semantic consistency.
Similarly, Unlu et al.~\cite{unlu2024groundup} have introduced GroundUp, a human-centered AI tool that enables architects to easily convert 2D sketches into 3D city massing models by integrating a sketch-to-depth prediction network with a diffusion model.
ControlCity~\cite{zhou2024controlcity} introduced a multimodal diffusion model that combines text, metadata, road networks, and imagery to accurately generate urban building footprint data from Volunteer Geographic Information (VGI). 
UrbanEvolver~\cite{qin2024urbanevolver} introduced a deep generative model for function-aware urban layout regeneration. The model generates roads and building layouts for target regions, considering land use types and surrounding contexts. It employs function-layout adaptive blocks and a comprehensive loss system.

\section{Summary and Discussion}

\label{discussions}
\subsection{Summary on existing methods}

In the literature on mainstream methodologies in computer-aided architectural design, it has been recognized that learning-based generation significantly transforms and enhances the traditional paradigm, which has been characterized by manually intensive, multi-round workflows for designers. Training on large-scale, high-quality benchmark datasets enables generation networks to learn desired layout distributions, even with adaptive user input or manually imposed constraints. The evolution of backbone neural networks from CNNs to Transformers, along with deep generative models such as GANs, VAEs, and Diffusion Models, has further improved generation performance across architectural design tasks. Another important shift in method design is the adoption of vectorized representations for layouts due to their flexibility and accuracy in depicting layout geometry. These learning-based generative paradigms offer greater flexibility, diversity, and completeness compared to traditional methods. However, significant limitations remain in these end-to-end synthesis pipelines, as discussed in the previous sections.

\begin{figure}[!htb]
    \centering
    \includegraphics[width=1.0\linewidth]{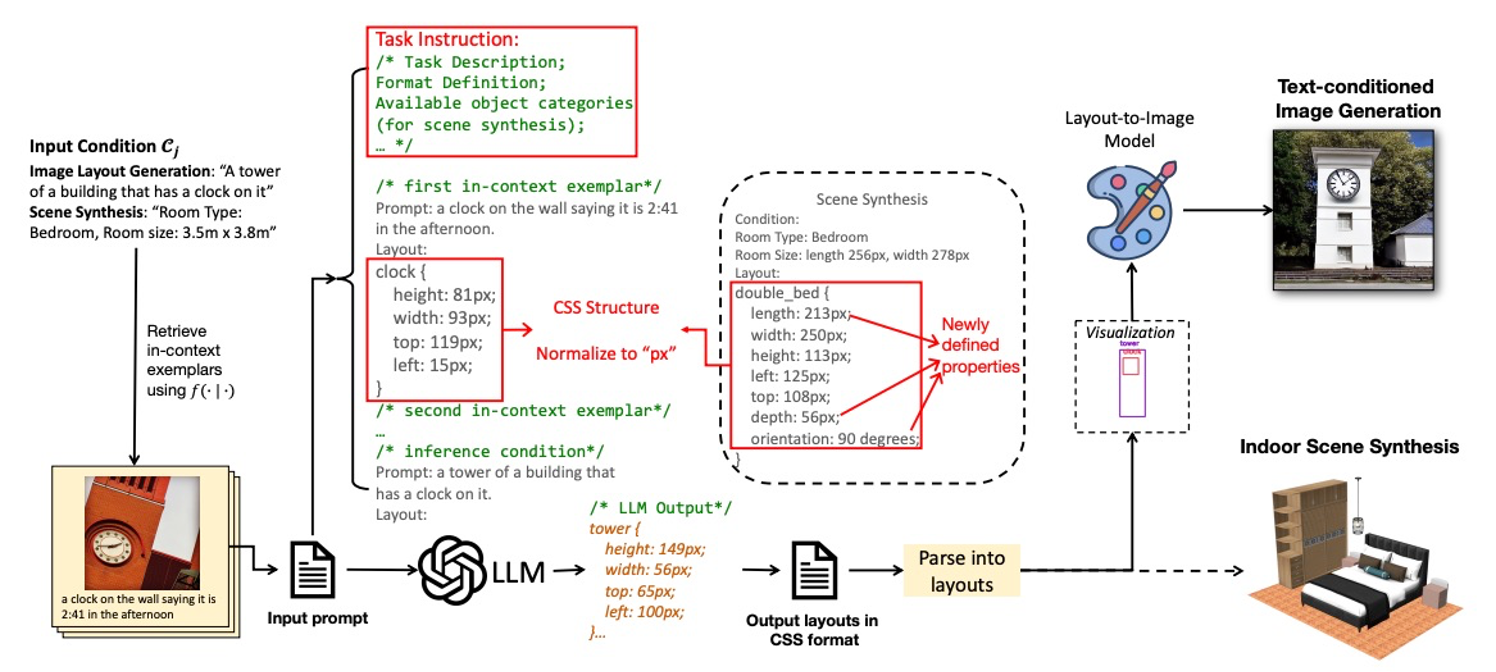}
    \caption{LayoutGPT achieves training-free layout-to-image or indoor scene layout generation using LLMs. With intricate text prompts and instructions~(\ie, converting the text inputs to CSS formats), it has demonstrated comparable or better generation fidelity with the SOTA methods trained on a specific dataset. }
    \label{fig::layoutgpt}
\end{figure}

\subsection{Discussion on open problems and new perspectives}

Future research could focus on multi-modal input support to accommodate diverse customer requirements. In floorplan generation, existing studies~\cite{wu2019data, nauata2020house} have explored user inputs such as house boundaries and bubble diagrams. In scene synthesis, current research emphasizes graph-conditioned~\cite{dhamo2021graph, zhai2024commonscenes, bahmani2023cc3d}, text-conditioned generation~\cite{tang2024diffuscene}, room boundary constraints~\cite{paschalidou2021atiss}, and partial layout completion~\cite{paschalidou2021atiss, para2023cofs, tang2024diffuscene}. Inspired by recent advancements in multi-modal, image-level foundational generative models~\cite{rombach2022high, ruan2023mm}, a promising direction is to develop a unified foundational floorplan generative model that accommodates various input formats (e.g., text or audio instructions, bubble diagrams, layout sketches), enabling the model to learn more powerful representations through the integration of multi-modal inputs. Moreover, reinforcement learning (RL) techniques can be applied to incorporate embodied AI into the layout generation process, allowing for better capture of user interactions and specifications.

Another potential direction is to leverage the capabilities of large language models (LLMs) to assist the architectural design workflow. Recent advancements in combining LLMs with visual understanding~\cite{pang2023frozen, yang2023dawn, wu2023visual} have inspired the community to explore how to fully utilize the potential of LLMs, such as ChatGPT, in supporting diverse downstream computer vision tasks. LayoutGPT~\cite{feng2024layoutgpt} (as shown in Fig.~\ref{fig::layoutgpt}) has demonstrated the remarkable efficacy of using LLMs (e.g., GPT-3.5 or GPT-4) for layout-to-image generation or indoor scene layouts. However, many state-of-the-art LLMs are proprietary and accessible only via paid APIs, limiting their broader adoption for handling large-scale data inputs. Furthermore, although LLMs are trained on extensive internet data to achieve superior zero-shot generative capabilities, they may not directly address the specialized needs of particular user inputs or tasks. As such, their output should be considered a preliminary result requiring further refinement. Furthermore, the generated layouts are generally limited to regular shapes, such as bounding boxes, which do not always align with the complex, often polygonal or non-Manhattan room layouts found in real-world floor plans. Further exploration is needed to adapt the broad applicability of LLMs to more specialized downstream tasks, such as floorplan generation and scene layout synthesis, using data from diverse environments with varied styles.


A shared limitation among most existing layout generative models is that they are trained and tested on a single dataset, severely limiting the generalizability of a model across different environments, object categories, scales, and styles. One potential solution is to construct large-scale benchmark datasets with more diverse data samples to enhance the model’s representation capacity and achieve desirable fidelity across varied data distributions. 
Another strategy to address dataset limitations is to employ self-supervised pre-training approaches, such as contrastive learning~\cite{chen2020simple, he2020momentum} or large foundation models~\cite{radford2021learning, ouyang2022training, rombach2022high}, to encapsulate a general embedded representation, which can then be fine-tuned for specific downstream tasks or datasets.

\section{Conclusions}

In this review paper, we first provide an overview of the current progress in computer-aided architectural design, categorizing the relevant topics into three major areas. For each category, we thoroughly examine methods from both the traditional architectural domain and learning-based generative approaches. We also discuss the advantages and current bottlenecks related to data, settings, and methods. Furthermore, we identify existing open problems and propose new perspectives for the research community, with a focus on leveraging powerful techniques such as foundational generative models and LLMs. We hope this survey encourages the community to reassess the needs and limitations of real-world applications, inspiring more valuable and insightful research in the field.



{\small
\bibliographystyle{cvm}
\bibliography{cvmbib}
}

\end{document}